\documentclass{article} 
\usepackage{iclr2026_conference,times}

\usepackage{amsmath,amsfonts,bm}

\def\eqref#1{equation~\ref{#1}}

\def\1{\bm{1}}

\DeclareMathAlphabet{\mathsfit}{\encodingdefault}{\sfdefault}{m}{sl}
\SetMathAlphabet{\mathsfit}{bold}{\encodingdefault}{\sfdefault}{bx}{n}

\usepackage{hyperref}
\usepackage{url}
\usepackage{times}
\usepackage{latexsym}
\usepackage[T1]{fontenc}
\usepackage[utf8]{inputenc}
\usepackage{microtype}
\usepackage{inconsolata}
\usepackage{graphicx}
\usepackage{booktabs}
\usepackage{multirow}
\usepackage{array}
\usepackage{xcolor}
\usepackage{amsmath}
\usepackage{listings}
\usepackage{wrapfig}

\newcommand{\customfont}{\fontsize{8.5pt}{9.5pt}\selectfont}

\title{Agentic Context Learning with Self-Discovered Specification}

\author{
\textbf{Jike Zhong}$^{1*\dagger}$,
\textbf{Ming Li}$^{2*}$,
\textbf{Yuxiang Lai}$^{3*}$,
\textbf{Ziyan Yang}$^{1}$,
\textbf{Jingyu Xie}$^{1}$,
\textbf{Jihyung Kil}$^{4}$, \\
\textbf{Zheda Mai}$^{5}$,
\textbf{Shao-Yuan Lo}$^{6}$,
\textbf{Xiang Ren}$^{1}$,
\textbf{Konstantinos Psounis}$^{1}$,
\textbf{Yuanyuan Lei}$^{2}$
\\[0.35em]
\normalsize
$^{1}$University of Southern California
\quad
$^{2}$University of Florida
\quad
$^{3}$Emory University
\\
$^{4}$Adobe Research
\quad
$^{5}$The Ohio State University
\quad
$^{6}$National Taiwan University
}

\iclrfinalcopy 
\begin{document}

\maketitle

\begingroup
\renewcommand{\thefootnote}{*}
\footnotetext{Equal contribution.}

\renewcommand{\thefootnote}{\ensuremath{\dagger}}
\footnotetext{Correspondence to: \texttt{jikezhon@usc.edu}.}
\endgroup

\begin{abstract}
Context learning is an emerging inference-time task where LLMs must learn and apply novel, task-specific knowledge from intricate contexts absent from pre-training; even frontier models score under 24\% task success. In this work, we conduct a comprehensive empirical study to understand why this setting remains difficult. A natural hypothesis is that failures stem from \textit{content access}; yet across twelve retrieval, reflection, and verification baselines on CL-Bench, an extensive context learning benchmark, we find limited gains over direct full-context prompting. Further failure analysis reveals a key finding: unlike typical long-context tasks such as long document understanding, context learning requires not only recovering local content but also acquiring \emph{local specifications} that are often \textit{unspecified} in the query but distributed across the context: domain-specific formats, local rules, and completeness conditions. Across all 31{,}592 rubric items, we find that 55.4\% clearly evaluate specification acquisition, while only 22.6\% evaluate content acquisition. Moreover, despite 76.7\% of specifications being unspecified in the user query, 95.5\% are traceable to the context, indicating these are learnable obligations rather than hidden requirements. 
To validate this diagnosis, we design a deliberately simple intervention PSCI (private specification-contract induction) which extracts local specifications and enforces them through adversarial checking and repair; PSCI achieves state-of-the-art 28.14\% with GPT-5.1 ($+5.59$ pp absolute and $+24.8\%$ relative) on CL-Bench, replicated on Qwen3.5-27B ($+5.28$ pp) and Gemini 3 Pro ($+6.17$ pp). Seventeen ablations further isolate the role of task-specific specifications. Overall, our results suggest context learning hinges on not only content acquisition but also specification acquisition.
\end{abstract}

\section{Introduction}
\label{sec:intro}

Large language models are increasingly deployed in settings where the prompt is not merely evidence to retrieve from, but defines a novel operating environment the model must learn before it can act: a product manual, workflow policy, regulatory code, API specification, or experimental dataset, often spanning tens of thousands of tokens and specifying the rules of a local domain. Consider the task of \textit{detecting anomalies in an incoming event stream}: the user prompt contains a 40K-token context followed by a simple query asking the model to record and log anomalies. The context defines not only the event semantics needed to detect anomalies, but also local specifications that determine what counts as a valid log: for instance, if a reserved event ID is skipped, the log must emit the exact flag \texttt{Sequence\_Gap\_Warning}. In this case, identifying the missing ID is the query-relevant content; emitting the prescribed flag is a validity-relevant specification. A response that detects the anomaly but omits the flag is still wrong. Thus, the challenge is not merely retrieving and reasoning over a long prompt, which is the focus of traditional long-context benchmarks~\citep{liu2024lost,hsieh2024ruler,bai2023longbench}, but acquiring the local specifications by which the answer is validated. 

\begin{figure*}[t]
  \centering
  \includegraphics[width=1\linewidth]{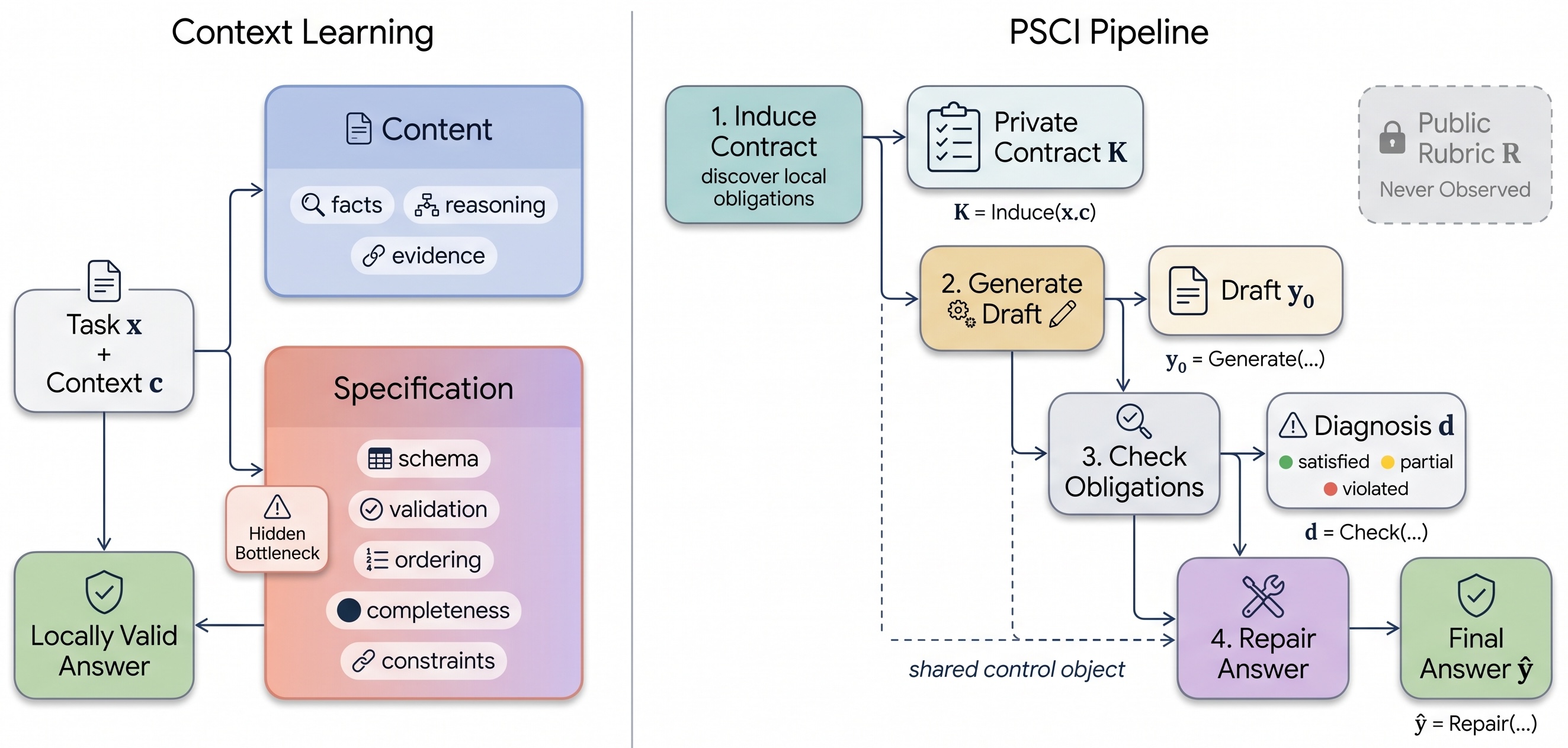}
\caption{\textbf{Left:} Context learning decomposes into content acquisition (what the answer says) and specification acquisition (how the answer must behave). \textbf{Right:} PSCI induces a private specification contract from the task and context, then uses it as a shared control object for generation, checking, and repair. The public rubric for the current task is never observed during any stage.}
  \label{fig:framework}
\vspace{-10pt}
\end{figure*}

Despite strong performance across many language tasks, frontier LLMs remain fragile on context learning: the best model in the original CL-Bench evaluation scores below 24\% strict task success. CL-Bench~\citep{dou2026clbench} is a recent benchmark designed specifically for context learning, with 1{,}899 diverse, long-context, contamination-controlled tasks (\autoref{sec:analysis}). In this paper, we conduct a comprehensive empirical study of why this setting remains difficult. A natural hypothesis is that failures stem from content access. To test this hypothesis, we evaluate twelve methods from four families on CL-Bench: retrieval and context restructuring~\citep{asai2024selfrag,yan2024corrective,sarthi2024raptor,edge2024graphrag,sun2023rankgpt}, test-time self-correction~\citep{shinn2023reflexion,chen2025guideline}, self-generated rubric verification~\citep{cook2024tick,wan2026inference,raghavendra2026agentic}, and iterative retrieval scaling~\citep{yue2025inference}. Surprisingly, none of the twelve exceeds the direct baseline (22.55\% strict pass) by more than one point (\autoref{sec:main}). This suggests that content access alone is insufficient and motivates a deeper investigation.

Analysis of failures reveals a clear pattern. Task-level strict pass is low (22.55\%), but rubric-level pass is much higher (76.01\%): models satisfy most requirements but miss a critical few, making failures near-misses rather than wholesale errors. Moreover, among tasks that fail by exactly one rubric item, 75.3\% of the missed items evaluate not factual content but compliance with locally specified behavioral constraints: formats, edge-case handling, ordering, prescribed labels, or completeness conditions. We call these obligations \textbf{local specifications} and the ability to discover and enforce them \textbf{specification acquisition}. Taxonomizing all 31{,}592 CL-Bench rubric items confirms this is systematic: 55.4\% evaluate specification acquisition, while only 22.6\% evaluate content acquisition.

A key asymmetry explains why direct context-to-answer generation misses specifications despite having access to the full context. Only 23.3\% of specification items are directly inferable from the user query; the remaining 76.7\% are recoverable \textit{only} from the surrounding context. Yet 95.5\% of specification items are traceable to the context itself through explicit declarations, validation patterns, or demonstrated examples. Crucially, this establishes them as \textbf{learnable obligations} distributed across the prompt rather than hidden evaluator preferences. Thus, the issue is not that local specifications are absent or incomprehensible. Rather, they are query-implicit, low-salience, and scattered across background conventions, schemas, exceptions, logging rules, and validation requirements. This also explains why retrieval-oriented methods give limited gains: they tend to surface query-relevant, answer-bearing evidence, while missing validity-relevant specifications that govern the response's form, procedure, validation behavior, or required action. A method may therefore find the \textit{content} needed to answer the apparent query while missing the local obligations needed for an acceptable response. 

Surprisingly, when explicitly prompted to \textit{induce specifications}, the model can recover much of this evaluation surface: privately induced contracts cover 94.4\% of held-out public specification rubrics with less than 0.5\% contradiction. This suggests that the missing step is not understanding the context in principle, but surfacing its local obligations as explicit generation constraints. To validate this diagnosis, we design a deliberately simple intervention: \textit{private specification-contract induction} (PSCI), a test-time scaffold comprising four steps (Figure~\ref{fig:framework}). The model first induces a task-specific specification contract from the context itself, then answers under the contract; an adversarial checker audits the draft against the contract; and a repair step patches violations. The public rubric for the current task is never observed at any stage. On the full 1{,}899-task benchmark, PSCI achieves state-of-the-art 28.14\% strict pass with GPT-5.1 (+5.59 pp absolute and +24.8\% relative), with consistent gains across all four task families and replication on Qwen3.5-27B (14.14\%$\!\to\!$19.42\%). \textbf{Seventeen} ablations isolate the role of task-specific specifications: shuffled contracts collapse below baseline, compute-matched generic critique returns to baseline, answer-first ordering fails. Overall, our contributions are threefold:
\begin{itemize}
\item Through comprehensive empirical analysis, we identify \emph{specification acquisition}---inferring query-implicit but context-traceable local obligations---as a key bottleneck in context learning, and show these obligations are recoverable by explicit induction.
\item To validate this diagnosis, we introduce PSCI, a deliberately simple test-time scaffold that achieves state-of-the-art CL-Bench performance across multiple model families.
\item Through extensive ablation and mechanism analysis, we show that specification acquisition must be grounded in the context and decomposes into two separable stages: specification discovery and specification enforcement.
\end{itemize}


\section{Related Work}
\label{sec:related}

\paragraph{Long-context evaluation and context learning.}
Long-context benchmarks evaluate evidence use over long inputs~\citep{kamradt2023niah,hsieh2024ruler,bai2023longbench,liu2024lost}, typically under a fixed task interface. Context learning additionally requires inducing local validity conditions from the prompt. CL-Bench~\citep{dou2026clbench} is the only large-scale benchmark explicitly designed for this setting. Unlike instruction-following benchmarks~\citep{zhou2023ifeval,xia2024fofo}, where constraints are explicit in the query, CL-Bench specifications are often query-implicit and must be induced from local documentation.

\paragraph{Retrieval, memory, and context restructuring.}
Methods improving content access~\citep{lewis2020rag,asai2024selfrag,yan2024corrective,sarthi2024raptor,edge2024graphrag,sun2023rankgpt,packer2023memgpt,yue2025inference} address a necessary but insufficient component of context learning: our twelve baselines show that content access alone does not solve specification acquisition (\autoref{sec:main}).

\paragraph{Critique, verification, and structured prompting.}
Self-refinement~\citep{madaan2023selfrefine,shinn2023reflexion}, checklist verification~\citep{cook2024tick,wan2026inference,raghavendra2026agentic}, and structured prompting~\citep{wei2022cot,yao2023react,khattab2024dspy} all introduce intermediate computation. PSCI differs in inducing a task-specific obligation set from context \emph{before} generation, then enforcing it through checking and repair. Ablations confirm this object, not the pipeline structure, drives the gain (\autoref{sec:ablations}).

Extended discussion in Appendix~\ref{app:related}.

\section{Empirical Study of Context Learning}
\label{sec:analysis}

\paragraph{Why CL-Bench.} We use CL-Bench~\citep{dou2026clbench} as our target because it is, to our knowledge, the only benchmark specifically designed for context learning: 1{,}899 tasks and 31{,}592 rubric items across four families (domain knowledge reasoning, rule system application, procedural task execution, and empirical discovery \& simulation), with contexts averaging 10.4K tokens and reaching 65K. All tasks are \textit{human-expert-written} and require \textit{knowledge absent from pre-training}, with contamination prevented by design (Appendix~\ref{app:clbench}). The breadth and quality of CL-Bench make it a strong testbed for an in-depth empirical study of context learning.

\paragraph{Initial hypothesis and experiment.} A natural hypothesis is that the low performance ($<24\%$ for GPT 5.1) stems from content access: the contexts are long and intricate, so methods that improve retrieval, restructuring, reflection, or verification should help. However, as shown in \autoref{sec:main}, \textbf{twelve} representative methods from these families provide limited gains over direct full-context prompting. This motivates a closer diagnosis of what models actually miss. To tackle this, we next conduct an in-depth analysis of the task itself.

\begin{table}[t]
\centering
\small
\begin{tabular}{lrr}
\toprule
Traceability class & Items & Share \\
\midrule
Explicit declaration & 12,531 & 71.6\% \\
Validation pattern & 3,693 & 21.1\% \\
Implicit example & 350 & 2.0\% \\
Convention / repeated struct. & 140 & 0.8\% \\
\textbf{Overall traceable} & \emph{16,714} & \emph{95.5\%} \\
\midrule
Snippet too short to judge & 718 & 4.1\% \\
Not derivable from context & 70 & 0.4\% \\
\bottomrule
\end{tabular}
\caption{Traceability of specification items to task/context. 95.5\% are traceable to explicit declarations, validation patterns, examples, repeated structures.}
\label{tab:traceability}
\vspace{-10pt}
\end{table}

\subsection{Analysis Protocol}
\label{sec:protocol}

We classify all 31{,}592 rubric items as evaluating \emph{content acquisition} or \emph{specification acquisition}: content if its verdict depends on which propositions the answer states; specification if it depends on whether the answer conforms to a locally defined rule governing form, procedure, or validation, or required action, holding propositions fixed. Two independent judges, an LLM-based classifier (GPT-5.1) and a deterministic rule-based classifier (with explicit abstention class), must agree for an item to be labeled; disagreements are marked \emph{ambiguous}.

To validate the taxonomy, we additionally audit a random sample of 300 rubric items with three PhD-level annotators. The human audit agrees with the conservative classifier labels in over 90\% of non-ambiguous cases (details in Appendix~\ref{app:protocol}).

\begin{wrapfigure}{r}{0.5\linewidth}
    \vspace{-30pt}
    \centering
    \includegraphics[width=1\linewidth]{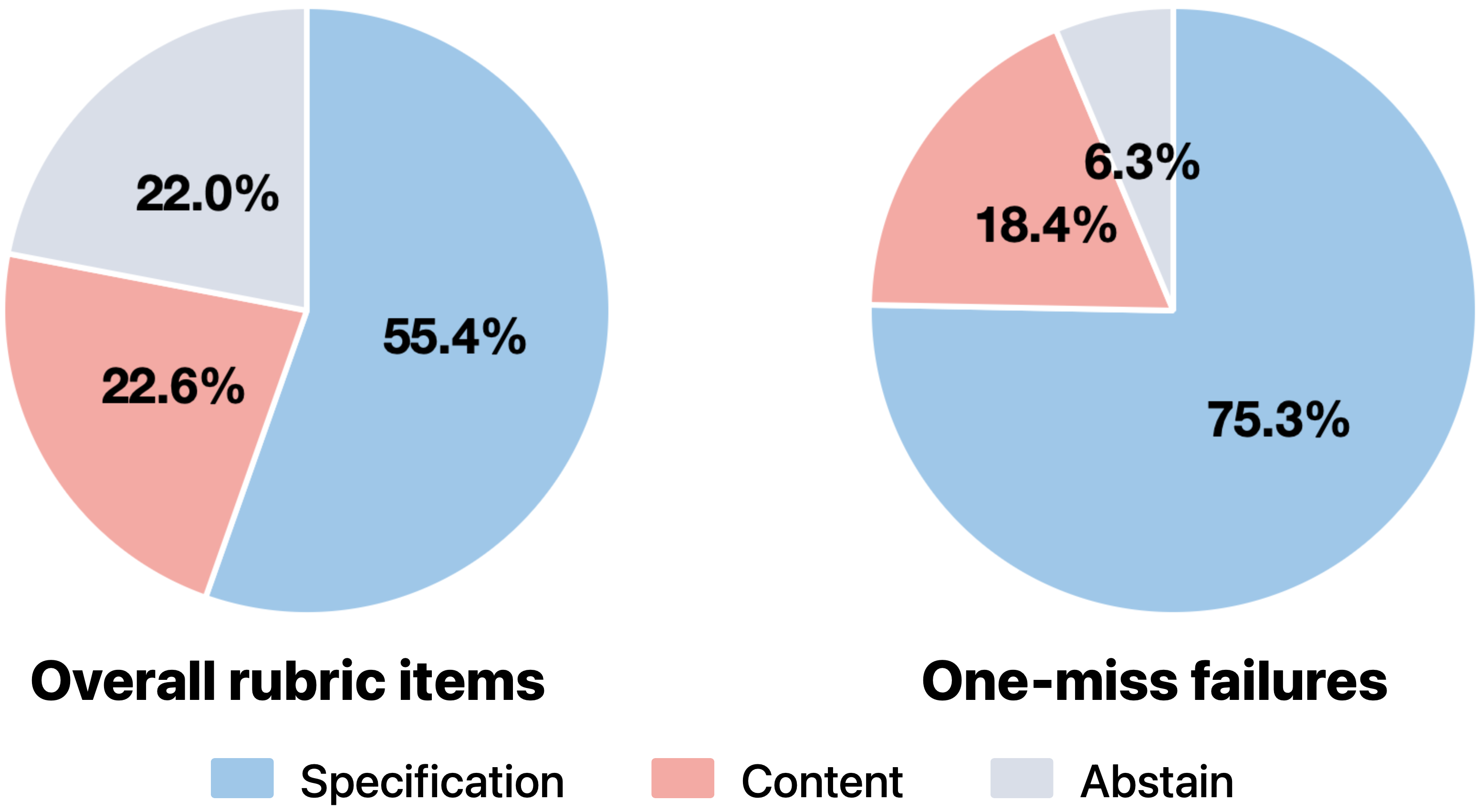}
    \caption{Rubric failure type breakdown by ``content acquisition'', ``specification acquisition'', and ``abstain''. Results show that the majority of original rubric items are ``specification acquisition," and it is also the main failure type among near miss tasks. }
    \label{fig:failure type}
    \vspace{-30pt}
\end{wrapfigure}

\subsection{Failures Are Often Near-Misses}
\label{sec:nearmiss}

The GPT-5.1 baseline (setup in \autoref{sec:setup}) passes 76.01\% of individual rubric items but only 22.55\% of tasks: most failures are near-misses rather than wholesale errors. Among the 281 tasks that fail by exactly one rubric item (validated in Appendix~\ref{app:nearmiss}), 75.3\% of the missed items are specification items---local conventions the model failed to enforce rather than missing facts. This concentration suggests that addressing specification acquisition would yield disproportionate gains.

\subsection{Specifications Form the Largest Share of Rubric Surface}
\label{sec:taxonomy}

Applying the protocol to all 31{,}592 items (\autoref{fig:failure type}), 55.4\% are classified as specification, 22.6\% as content, and 22.0\% as ambiguous. Thus, even under a conservative criterion, specification acquisition accounts for the largest share of CL-Bench's evaluation surface. The specification class further decomposes into eight subtypes---from validation checks (42.9\%) to role/tone constraints (1.2\%)---confirming it captures multiple dimensions of locally valid behavior rather than surface formatting alone (full breakdown in Table~\ref{tab:subtypes}).

\begin{table}[t]
\centering
\footnotesize
\setlength{\tabcolsep}{4pt}
\begin{tabular}{lrrrr}
\toprule
Specification subtype & Items & Share & Base & PSCI \\
\midrule
validation\_or\_edge\_case & 7,508 & 42.9\% & 81.0 & 85.0 \\
exclusion\_or\_forbidden & 2,573 & 14.7\% & 80.1 & 85.3 \\
ordering\_or\_sequence & 2,205 & 12.6\% & 68.4 & 74.6 \\
schema\_or\_structured & 1,663 & 9.5\% & 78.4 & 82.9 \\
source\_grounding & 1,085 & 6.2\% & 79.1 & 84.1 \\
coverage\_or\_completeness & 805 & 4.6\% & 65.9 & 75.7 \\
exact\_phrase\_constraint & 525 & 3.0\% & 66.4 & 73.3 \\
role\_tone\_style & 210 & 1.2\% & 59.1 & 72.0 \\
\bottomrule
\end{tabular}
\caption{Specification subtype distribution and item-level pass rates by baseline and PSCI (\%). Results show that every subtype is improved by PSCI; the largest gains are concentrated in harder subtypes.}
\label{tab:subtypes}
\vspace{-10pt}
\end{table}

\subsection{Specifications Are Learnable but Query-Implicit}
\label{sec:traceability}

A natural concern is that specification items encode arbitrary evaluator preferences. We test this directly (Table~\ref{tab:traceability}; annotation protocol in Appendix~\ref{app:traceability}): 95.5\% of specification items are traceable to the context through explicit declarations, validation patterns, implicit examples, or repeated conventions. Only 0.4\% are not derivable.

However, only 23.3\% of specification items are inferable from the user query alone; the remaining 76.7\% require reading the surrounding context. This asymmetry explains why both direct prompting and retrieval methods underperform: specifications are learnable obligations distributed across the prompt, but they are query-implicit, low-salience, and scattered across background conventions. Crucially, the problem is not that specifications are incomprehensible---when explicitly prompted to extract them, the model recovers 94.4\% of held-out public rubrics with less than 0.5\% contradiction. The capability exists, but is not reliably exercised during standard generation.

\section{Private Specification-Contract Induction}
\label{sec:method}

The analysis in \autoref{sec:analysis} suggests that context learning failures are not explained by content access alone: many missed requirements are local specifications that are traceable to the context but largely absent from the user query. To validate this diagnosis, we design a deliberately simple intervention, Private Specification-Contract Induction (PSCI). PSCI tests whether making these latent specifications explicit before generation and enforcing them when forming responses improves context learning: 
\begin{align}
K &= \mathrm{Induce}(x, c;\, \theta) \label{eq:induce} \\
y_0 &= \mathrm{Generate}(x, c, K;\, \theta) \label{eq:generate} \\
d &= \mathrm{Check}(x, c, K, y_0;\, \theta) \label{eq:check} \\
\hat{y} &= \mathrm{Repair}(x, c, K, y_0, d;\, \theta) \label{eq:repair}
\end{align}
where $x$ is the task, $c$ is the context, $K$ is the induced specification contract, $y_0$ is the initial draft, $d$ is a structured diagnosis, $\hat{y}$ is the final answer, and $\theta$ denotes the LLM parameters. The public rubric for the current task is never observed at any stage. The method is deliberately minimal: a complex pipeline would obscure whether gains stem from the diagnosis being correct or from engineering.

\subsection{Specification Discovery}
\label{sec:discovery}

Given task $x$ and context $c$, an LLM induces a private specification contract $K = \{k_1, k_2, \ldots, k_n\}$: a structured set of $n$ obligation items the answer must satisfy to be locally valid, covering required formats, validation checks, ordering constraints, and exact local artifacts. The model then produces a draft $y_0 = \mathrm{Generate}(x, c, K;\, \theta)$ conditioned on the task, context, and contract jointly. We use rotated demonstrations drawn from disjoint context learning tasks only to teach contract style while the final contract used for generation is induced from the current task context; no same-task answer, public rubric, or task-specific contract is shown at any stage (details in Appendix~\ref{app:exemplars}). \textbf{For analysis only}, after generation, we measure specification discovery quality via contract--rubric overlap:
\begin{equation}
\begin{split}
\mathrm{Coverage}(K, R) &= \\
&\displaystyle \frac{|\{r \in R : \exists\, k \in K \text{ that covers } r\}|}{|R|}
\end{split}
\label{eq:coverage}
\end{equation}

\subsection{Specification Enforcement}
\label{sec:enforcement}

An adversarial checker functions as a diagnostic verifier. Given $(x, c, K, y_0)$, it assesses each obligation $k_i \in K$ and produces:
\begin{equation}
\begin{split}
d = \{&(k_i,\, s_i,\, r_i)\}_{i=1}^{n}, \\
&s_i \in \{\textsc{Satisfied},\, \textsc{Partial},\, \textsc{Violated}\}
\end{split}
\label{eq:diagnosis}
\end{equation}
where $s_i$ is the compliance status and $r_i$ is a natural-language rationale localizing the mismatch. The repair model produces $\hat{y} = \mathrm{Repair}(x, c, K, y_0, d;\, \theta)$, preserving correct content while patching obligations diagnosed as \textsc{Partial} or \textsc{Violated}. Crucially, the checker targets the induced contract $K$, not the draft's surface quality---decoupling verification from generation. Two design decisions distinguish PSCI from generic self-refinement: the contract is induced \emph{before} generation (control object, not post-hoc grader), and the checker is grounded in $K$ rather than open-ended quality. Both are verified empirically in \autoref{sec:ablations} (more analysis in Appendix~\ref{app:design}).

\begin{table*}[t]
\centering
\customfont
\scriptsize
\setlength{\tabcolsep}{1pt}
\begin{tabular}{ll*{11}{c}}
\toprule
& & \multicolumn{3}{c}{Overall} & \multicolumn{2}{c}{DKR} & \multicolumn{2}{c}{RSA} & \multicolumn{2}{c}{PTE} & \multicolumn{2}{c}{EDS} \\
\cmidrule(lr){3-5} \cmidrule(lr){6-7} \cmidrule(lr){8-9} \cmidrule(lr){10-11} \cmidrule(lr){12-13}
Model & Method & Strict & Rubric & Task & Strict & Rubric & Strict & Rubric & Strict & Rubric & Strict & Rubric \\
\midrule
\multirow{10}{*}{\shortstack{\textit{Other frontier}\\ \textit{models$^*$}}}
 & GPT 5.1 (High) & 23.7$\pm$0.5 & -- & -- & 25.3 & -- & 23.7 & -- & 23.8 & -- & 18.1 & -- \\
 & Claude Opus 4.5 Think. & 21.1$\pm$1.4 & -- & -- & 23.7 & -- & 19.0 & -- & 22.6 & -- & 15.1 & -- \\
 & GPT 5.2 (High) & 18.1$\pm$0.8 & -- & -- & 18.6 & -- & 17.2 & -- & 21.4 & -- & 11.7 & -- \\
 & o3 (High) & 17.8$\pm$0.2 & -- & -- & 18.0 & -- & 17.6 & -- & 19.5 & -- & 13.7 & -- \\
 & Kimi K2 Think. & 17.6$\pm$0.6 & -- & -- & 18.7 & -- & 17.0 & -- & 18.8 & -- & 12.6 & -- \\
 & HY 2.0 Think. & 17.2$\pm$0.6 & -- & -- & 18.0 & -- & 17.3 & -- & 19.4 & -- & 8.9 & -- \\
 & Gemini 3 Pro (High) & 15.8$\pm$0.3 & -- & -- & 15.5 & -- & 17.7 & -- & 16.4 & -- & 10.1 & -- \\
 & Qwen 3 Max Think. & 14.1$\pm$0.1 & -- & -- & 13.5 & -- & 15.6 & -- & 15.2 & -- & 9.0 & -- \\
 & Doubao 1.6 Think. & 13.4$\pm$0.1 & -- & -- & 13.7 & -- & 14.2 & -- & 13.9 & -- & 9.4 & -- \\
 & DeepSeek V3.2 Think. & 13.2$\pm$0.4 & -- & -- & 13.6 & -- & 13.8 & -- & 14.2 & -- & 8.0 & -- \\
\midrule
\multirow{18}{*}{\shortstack{\textit{GPT-5.1}\\ \textit{(Medium)}}} 
 & Baseline & 22.55$\pm$0.6 & 76.01$\pm$1.2 & 71.87 & 23.83 & 80.23 & 23.23 & 72.97 & 21.06 & 75.29 & 18.18 & 69.87 \\
\addlinespace[4pt]
 & Self-RAG & 20.67$\pm$0.7 & 71.11$\pm$1.8 & 66.55 & 24.07 & 75.13 & 17.98 & 68.33 & 21.43 & 73.77 & 15.15 & 61.61 \\
 & CRAG & 19.00$\pm$1.4 & 70.79$\pm$2.1 & 68.13 & 24.07 & 77.07 & 13.48 & 64.30 & 20.00 & 77.68 & 6.06 & 54.82 \\
 & RankGPT & 22.33$\pm$2.4 & 69.72$\pm$2.5 & 67.37 & 22.22 & 76.42 & 10.11 & 63.90 & 22.86 & 76.27 & 18.18 & 52.12 \\
 & RAPTOR & 20.67$\pm$0.2 & 71.30$\pm$1.4 & 66.96 & 28.70 & 79.59 & 16.85 & 69.35 & 17.14 & 75.25 & 12.12 & 47.59 \\
 & GraphRAG & 23.33$\pm$0.5 & 75.47$\pm$1.9 & 67.57 & 28.63 & 77.82 & 14.61 & 65.09 & \textbf{35.71} & 80.03 & 9.09 & 75.35 \\
 & SimpleMem & 20.00$\pm$1.5 & 73.19$\pm$2.0 & 67.00 & 29.63 & 78.59 & 12.36 & 67.14 & 21.43 & 75.41 & 6.06 & 67.00 \\
\addlinespace[4pt]
 & Reflexion & 22.33$\pm$1.2 & 71.64$\pm$1.9 & 70.79 & 26.85 & 79.82 & 19.10 & 69.12 & 20.00 & 72.83 & \textbf{21.21} & 54.25 \\
 & Guideline Forest & 22.00$\pm$0.9 & 73.82$\pm$1.8 & 70.19 & 26.85 & 81.35 & 16.85 & 70.22 & 22.86 & 79.09 & 18.18 & 52.55 \\
\addlinespace[4pt]
 & TICK/STICK & 23.00$\pm$0.3 & 74.20$\pm$1.4 & 67.40 & 27.70 & 76.30 & 20.22 & 67.85 & 27.14 & 74.16 & 12.12 & 78.74 \\
 & DeepVerifier & 21.33$\pm$2.7 & 70.55$\pm$2.2 & 65.90 & 29.63 & 79.71 & 16.85 & 60.82 & 17.14 & 68.52 & 15.15 & 69.55 \\
 & Agentic Rubrics & 20.33$\pm$2.2 & 69.99$\pm$1.7 & 65.10 & 26.85 & 72.67 & 14.61 & 65.64 & 24.29 & 69.22 & 6.06 & 72.80 \\
\addlinespace[4pt]
 & IterDRAG & 22.00$\pm$0.1 & 72.10$\pm$2.9 & 66.87 & 28.70 & 73.67 & 20.22 & 65.88 & 18.57 & 74.31 & 12.12 & 75.50 \\
\cmidrule(lr){2-13}
 & Baseline+SI & 26.09$\pm$1.2$_{+3.51}$ & 79.10$\pm$1.6$_{+3.09}$ & 75.39 & 28.96 & 84.21 & 24.29 & 74.52 & 27.23 & 78.83 & \textbf{21.21} & 72.48 \\
 & PSCI & \textbf{28.14}$\pm$0.5$_{+5.59}$ & \textbf{81.13}$\pm$0.9$_{+5.12}$ & \textbf{77.57} & \textbf{31.67} & \textbf{84.84} & \textbf{25.18} & \textbf{76.70} & 29.15 & \textbf{81.04} & \textbf{21.21} & \textbf{79.12} \\
 \midrule
\multirow{3}{*}{\textit{Gemini 3 Pro}}
 & Baseline & 16.14$\pm$0.2 & 70.31$\pm$1.2 & 68.64 & 15.60 & 69.55 & 18.96 & 62.48 & 16.82 & 68.49 & 12.12 & 57.33 \\
 & Baseline+SI & 18.38$\pm$0.6$_{+2.24}$ & 71.71$\pm$1.4$_{+1.40}$ & 70.89 & 19.70 & 73.34 & 17.89 & 72.45 & 18.43 & 72.95 & 14.77 & 64.27 \\
 & PSCI & \textbf{22.31}$\pm$0.1$_{+6.17}$ & \textbf{76.75}$\pm$2.6$_{+5.44}$ & \textbf{72.02} & \textbf{23.42} & \textbf{79.16} & \textbf{23.05} & \textbf{74.54} & \textbf{20.40} & \textbf{75.61} & \textbf{17.91} & \textbf{68.83 } \\
\midrule
\multirow{3}{*}{\textit{Qwen3.5-27B}}
 & Baseline & 14.14$\pm$2.7 & 69.41$\pm$2.9 & 63.87 & 15.84 & 72.97 & 12.06 & 65.49 & 17.02 & 71.58 & 8.59 & 54.61 \\
 & Baseline+SI & 17.41$\pm$1.8$_{+3.27}$ & 73.10$\pm$4.0$_{+3.69}$ & 69.00 & 18.85 & 76.59 & 17.02 & 68.44 & 19.15 & 73.97 & 10.61 & 65.36 \\
 & PSCI & \textbf{19.42}$\pm$2.4$_{+5.28}$ & \textbf{74.58}$\pm$3.3$_{+5.17}$ & \textbf{70.98} & \textbf{20.97} & \textbf{77.44} & \textbf{18.79} & \textbf{71.20} & \textbf{21.49} & \textbf{75.40} & \textbf{12.12} & \textbf{65.65} \\
\bottomrule
\end{tabular}
\caption{Main results on CL-Bench (1{,}899 tasks, 31{,}592 rubric items). \emph{Overall}: strict-pass rate, global rubric (item micro-average), and mean task rubric (task macro-average). Per-family columns: strict-pass and rubric for DKR (Domain Knowledge Reasoning), RSA (Rule System Application), PTE (Procedural Task Execution), EDS (Empirical Discovery \& Simulation). Baselines grouped by family (a--d) as defined in \autoref{sec:setup}. Subscripts = absolute gains over same-model baseline; best in \textbf{bold}. PSCI improves all evaluated model groups, surpassing even GPT-5.1 High (23.7\%). $^*$Frontier results from \citet{dou2026clbench}; rubric/task columns unavailable.}
\label{tab:main}
\vspace{-10pt}
\end{table*}

\section{Experiments}
\label{sec:experiments}

\subsection{Setup}
\label{sec:setup}

\paragraph{Benchmark.}
We evaluate on CL-Bench~\citep{dou2026clbench}, which contains 1{,}899 tasks across four families: domain knowledge reasoning (DKR), rule system application (RSA), procedural task execution (PTE), and empirical discovery \& simulation (EDS). The benchmark is scored by 31{,}592 rubric items, averaging 16.64 items per task. Contexts average 10.4K tokens and reach 65K. A task is solved only when \emph{all} associated rubric items pass; even GPT-5.1 with high reasoning effort achieves only 23.7\% strict pass in the original evaluation.

\paragraph{Models.}
We use GPT-5.1~\citep{singh2025gpt5} with medium reasoning effort as the primary model. GPT-5.1 is the strongest model in the original CL-Bench evaluation, and medium effort gives the strongest direct baseline in our setup, making it a conservative comparison point for PSCI. To test cross-model generalization, we additionally evaluate on Gemini 3 Pro (high) \citep{GeminiTeam2025} and Qwen3.5-27B~\citep{yang2025qwen3}.

\paragraph{Baselines.}
The direct baseline is single-turn full-context prompting. We compare against twelve methods from four families: \textbf{(a)} retrieval and context restructuring: Self-RAG~\citep{asai2024selfrag}, CRAG~\citep{yan2024corrective}, RAPTOR~\citep{sarthi2024raptor}, GraphRAG~\citep{edge2024graphrag}, RankGPT~\citep{sun2023rankgpt}, and SimpleMem; \textbf{(b)} self-correction: Reflexion~\citep{shinn2023reflexion} and Guideline Forest~\citep{chen2025guideline}; \textbf{(c)} self-generated rubric verification: TICK/STICK~\citep{cook2024tick}, DeepVerifier~\citep{wan2026inference}, and Agentic Rubrics~\citep{raghavendra2026agentic}; and \textbf{(d)} iterative retrieval: IterDRAG~\citep{yue2025inference}. All baselines use the same GPT-5.1 medium-effort setting with full prompts and implementation details provided in \autoref{app:eval}.

\begin{table*}[t]
\centering
\customfont
\setlength{\tabcolsep}{5pt}
\begin{tabular}{llcc}
\toprule
& Method & Strict Pass (\%) & Rubric Pass (\%) \\
\midrule
\multirow{8}{*}{\shortstack[l]{Pipeline\\ablations}}
 & Baseline & 22.55$\pm$0.6 & 76.01$\pm$1.2 \\
 & A1: + Spec & 26.09$\pm$1.2 & 79.10$\pm$1.6 \\
 & A2: + Repair only & 23.02$\pm$0.4 & \emph{71.68}$\pm$1.6 \\
 & A3: + Chk + Repair (no spec) & 24.32$\pm$0.7 & 77.44$\pm$1.5 \\
 & A4: + Spec + Repair (no chk) & 26.30$\pm$1.0 & 80.15$\pm$2.3 \\
 & A5: + Spec + Combined Chk\&Rep & 25.90$\pm$0.4 & 78.80$\pm$2.5 \\
 & A6: + Answer-first Spec + Chk + Rep & 23.67$\pm$1.0 &\emph{70.67}$\pm$1.7 \\
 & \textbf{PSCI} (Ours) & \textbf{28.14}$\pm$0.5 & \textbf{81.13}$\pm$0.9 \\
\midrule
\multirow{3}{*}{\shortstack[l]{Spec quality\\(discovery only)}}
 & A7: + Spec (Shuffled) & 19.80$\pm$1.5 & \emph{66.55}$\pm$1.2 \\
 & A8: + Spec (Zero-shot) & 25.01$\pm$0.7 & \emph{72.69}$\pm$2.1 \\
 & A9: + Spec (Generic) & 23.80$\pm$0.9 & 76.76$\pm$2.6 \\
\midrule
\multirow{3}{*}{\shortstack[l]{Spec quality\\(full pipeline)}}
 & A10: + Spec (Shuffled) + Chk + Rep & 22.00$\pm$1.2 & \emph{66.55}$\pm$2.1 \\
 & A11: + Spec (Zero-shot) + Chk + Rep & 26.67$\pm$1.1 & \emph{78.69}$\pm$1.2 \\
 & A12: + Spec (Generic) + Chk + Rep & 24.00$\pm$0.5 & 74.76$\pm$1.7 \\
\midrule
\multirow{5}{*}{\shortstack[l]{Compute-\\matched\\controls}}
 & A13: + Spec-free Critique & 22.67$\pm$0.8 & \emph{70.06}$\pm$0.9 \\
 & A14: + Self-reflection & 23.02$\pm$0.4 & \emph{71.24}$\pm$1.4 \\
 & A15: + Majority Vote & 24.12$\pm$0.9 & \emph{74.23}$\pm$1.9 \\
 & A16: + Repeated Static Spec & 24.30$\pm$1.9 & 78.45$\pm$3.1 \\
 & A17: + Repeated Zero-shot Spec & 24.67$\pm$1.8 & 79.10$\pm$1.5 \\
\bottomrule
\end{tabular}
\caption{Ablation study (full benchmark, GPT-5.1 Medium). ``+'' = added to Baseline. Spec = specification contract; Chk = checker; Rep = repair. \emph{Italics} = rubric below baseline (76.01\%). We highlight three key results: shuffled specs collapse below baseline (A7/A10), spec-free critique returns to baseline (A13)—extra compute alone does not explain the gain, and answer-first ordering underperforms PSCI by 4.47 pp.}
\label{tab:ablation}
\vspace{-10pt}
\end{table*}

\paragraph{Metrics and protocol.}
We report strict task pass rate, global rubric pass rate (item-level micro-average), and mean task rubric (task-level macro-average), both overall and by task family. We follow the released CL-Bench evaluation protocol, using the same GPT-5.1 verifier and evaluation code, which the benchmark reports as achieving over 90\% agreement with human judgments. For each method and model, we run the full benchmark three times and report mean $\pm$ standard deviation. The complete PSCI pipeline constitutes one run; reruns are used only for API infrastructure failures, with no selective filtering or cherry-picking (additional eval info in \autoref{app:eval}). 

To validate that gains are not an artifact of automatic evaluation, three PhD-level annotators blindly audit 100 baseline-fail/PSCI-pass tasks. A majority vote prefers the PSCI answer in 96/100 (96\%) of cases and confirms the automatic flip judgment in 92/100 (92\%) (protocol in \autoref{app:human_flips}).

\subsection{Main Results}
\label{sec:main}

\paragraph{PSCI improves across model families.}
Table~\ref{tab:main} reports full-benchmark results. PSCI improves GPT-5.1 strict pass from 22.55\% to 28.14\%, a +5.59 pp absolute gain and 24.8\% relative improvement. Global rubric pass also increases by 5.12 pp. The improvement replicates on Qwen3.5-27B, where PSCI improves strict pass from 14.14\% to 19.42\% (+5.28 pp) and global rubric pass by +5.17 pp. The gain also holds on Gemini 3 Pro, improving strict pass from 16.14\% to 22.31\% (+6.17 pp). Gains are consistent across all four CL-Bench task families. Notably, PSCI with GPT-5.1 medium (28.14\%) exceeds GPT-5.1 high from the original CL-Bench (23.7\%), suggesting explicit specification induction is a more effective use of inference compute than increasing reasoning effort alone.

\paragraph{The gains align with the diagnosis.}
PSCI improves both rubric types, but gains are larger on specifications. On GPT-5.1, specification-item pass rate rises from 76.7\% to 82.1\%, closing 23.2\% of the available gap; content-item pass rate rises from 73.4\% to 77.5\%, closing 15.5\%. Among rubric items fixed on tasks that flip from baseline-fail to PSCI-pass, 77.7\% are specification items, showing PSCI primarily addresses the diagnosed specification bottleneck while also improving content use.

\paragraph{Existing methods do not match PSCI.}
As discussed in \autoref{sec:intro}, all twelve alternative methods remain close to the direct baseline, ranging from 19.00\% to 23.33\% strict pass, with none exceeding the baseline by more than 0.78 percentage points. Retrieval and restructuring methods often underperform full-context prompting, consistent with the hypothesis that context selection can discard distributed specification signals. Self-generated rubric methods, which are closest in spirit to PSCI, also fail to meaningfully exceed the baseline. Therefore, the gain is not explained by generic checklist verification; it requires context-grounded specification induction and obligation-level checking.

\subsection{Ablation Studies}
\label{sec:ablations}

To isolate the mechanism, we run 17 ablations (A1--A17) on the full 1{,}899-task benchmark under the same GPT-5.1 medium-effort setting (Table~\ref{tab:ablation}).

\paragraph{Each PSCI component contributes.}
Specification induction alone (A1) reaches 26.09\%, accounting for 63\% of PSCI's total strict-pass gain. Checker and repair add a further +2.05 points. In contrast, repair alone (A2, 23.02\%) and checker+repair without a specification contract (A3, 24.32\%) yield only limited gains, showing that the contract is the primary driver. Separating diagnosis from repair is also important: spec+repair without an explicit checker (A4, 26.30\%) and a combined check-and-repair step (A5, 25.90\%) both underperform PSCI.

\paragraph{Extra compute does not explain the gain.}
Five compute-matched alternatives without task-specific specifications all fall short. Spec-free critique (A13) returns to 22.67\%, and self-reflection (A14) to 23.02\%. Even repeated contract-like structures without task-specific grounding (A16--A17) plateau at 24.30--24.67\%. Thus, additional inference and generic self-correction are insufficient; the gain requires a task-specific specification contract.

\paragraph{Finding 1: Specifications must be extracted separately (A7, A10).}
To test whether contracts are merely helpful prompt prefixes, we replace each contract with one drawn from another task. Shuffled contracts collapse below baseline: A7 drops to 19.80\% strict pass and A10 to 22.00\%, with rubric pass falling to 66.55\%. A misaligned specification is worse than no specification, demonstrating that task-specific alignment is causally necessary.

\begin{table}[t]
\centering
\customfont
\begin{tabular}{lrrrr}
\toprule
& Items & Strong/Human & Partial & Contradict  \\
\midrule
Content     & 7{,}140  & 54.9/60.4 & 38.7 & 0.38 \\
Spec  & 17{,}501 & 62.5/69.5 & 31.9 & 0.52 \\
All         & 24{,}640 & 60.8/64.95 & 33.4 & 0.49 \\
\bottomrule
\end{tabular}
\caption{Overlap (\%) between privately induced contracts and non-ambiguous public (ground-truth) rubrics. The contract recovers most of the public rubric surface, especially for specification items.}
\label{tab:overlap}
\vspace{-15pt}
\end{table}

\paragraph{Finding 2: Specifications must be grounded in context (A8, A11, A12).}

Generic specifications underperform context-derived ones. Zero-shot context-derived specifications through the full pipeline (A11) reach 26.67\%, while generic static specifications (A12) reach only 24.00\%. The hierarchy is clear: generic specifications $\ll$ zero-shot context-derived specifications $<$ few-shot context-derived PSCI. Zero-shot context-derived specifications already improve over baseline, while generic and shuffled specifications underperform, showing that demonstrations refine contract granularity but do not create the effect. Extracting task-specific specifications from the current context is key.



\paragraph{Finding 3: Specifications must be enforced (A1, A4--A5).}
Discovery alone (A1, 26.09\%) remains below PSCI, showing that specification discovery is not sufficient. Spec+repair without an explicit checker (A4, 26.30\%) and combined check-and-repair (A5, 25.90\%) also underperform PSCI, indicating that enforcement benefits from explicit checking and targeted repair. This separates discovery from compliance: the checker exposes concrete spec violation targets, enabling compliant repair rather than unconstrained rewriting.


\paragraph{Finding 4: Contracts Must Act as Control Objects.}
\label{sec:ordering}

The answer-first ablation (A6) tests whether specifications can be induced after generation and used only as repair guidance. It cannot: A6 drops to 23.67\% strict pass and 70.67\% rubric pass, far below PSCI. Since only 23.3\% of specifications are inferable from the user query, direct generation often fixes the answer structure before scattered obligations become active constraints. Post-hoc repair can patch local violations, but cannot reliably restructure fields, evidence use, or completeness conditions. Thus, the contract must shape generation as a control object rather than act as a post-hoc grader, explaining why self-generated rubric verification methods do not match PSCI.

\section{Discussion}
\label{sec:discussion}

\subsection{Quality of Induced Specifications}
\label{sec:overlap}

For analysis only, we compare private contracts with held-out public rubrics never observed by PSCI, labeling each item as strongly covered, partially covered, contradicted, or uncovered (details in \autoref{app:coverage}). Across 24{,}640 non-ambiguous items, contracts strongly cover 60.8\% and partially cover 33.4\%, totaling 94.2\% coverage with only 0.49\% contradiction (Table~\ref{tab:overlap}). Specification items are more strongly covered than content items (62.5\% vs.\ 54.9\%), and human audit confirms the automatic measurement is conservative (64.95\% vs.\ 60.8\% strong coverage; Appendix~\ref{app:human_overlap}).

Coverage also tracks correctness: pass rate rises from contradicted (25.4\%) and uncovered (48.7\%) items to partially covered (75.2\%) and strongly covered items (88.8\%). Thus, induced contracts are not decorative; better coverage corresponds to downstream success, with a 40-point pass-rate gap between strongly covered and uncovered items.

\subsection{The Enforcement Bottleneck}
\label{sec:residual}

Among rubric items still failed by PSCI, 83\% are covered by the contract but not executed in the final answer; only 14.9\% are discovery failures and 2.0\% are contradictions. Thus, PSCI substantially improves specification discovery, but enforcement becomes the dominant residual bottleneck: current models often surface the right local obligations but fail to consistently satisfy them during generation and repair. Future work should prioritize stronger enforcement mechanisms, such as iterative verification, more reliable repair, or training-time objectives that internalize specification compliance.

\section{Conclusion}
\label{sec:conclusion}

We identify local specification acquisition as a key bottleneck in context learning. This motivates PSCI, a simple test-time scaffold that makes local specifications explicit before generation and enforces them through checking and repair. Our results show that context learning requires not only content acquisition, but also context-grounded specification discovery and enforcement.

\clearpage
\section*{Acknowledgment}
This work is partially supported by a grant from the USC-Amazon Center on Secure \& Trusted ML and funding from the National Science Foundation NSF (Award \# (USC): 1956435).

\section*{Limitations}
\label{sec:limitations}
\paragraph{Benchmark scope.}
We mainly target our analysis and experiments on the single benchmark, CL-Bench, designed specifically for context learning. Although this limits claims about generalization beyond the current context learning benchmark landscape on the surface, CL-Bench is currently the only large-scale benchmark explicitly designed for context learning, and it is unusually broad for a single benchmark: 1{,}899 human-expert-written, contamination-controlled tasks, 31{,}592 rubric items, four task families, and contexts averaging 10.4K tokens and reaching 65K. Thus, while evaluation on future context learning benchmarks would strengthen external validity, CL-Bench already covers a diverse set of local domains, including domain knowledge, rule systems, procedures, and empirical discovery.

\paragraph{Inference cost.}
PSCI is an inference-time scaling method and is therefore more expensive than single-pass generation: it uses four sequential model calls for specification induction, contract-guided answer generation, contract-aware checking, and targeted repair. We treat this as a central tradeoff rather than a hidden cost. Compute-matched and compute-heavier controls show that additional inference alone does not explain the gain: generic self-reflection, iterative retrieval, guideline-based reasoning, and verification-style baselines use comparable or larger budgets but fail to match PSCI. Thus, the improvement comes from how computation is structured---around task-specific specification discovery and enforcement---rather than from extra tokens alone. Future work should explore cheaper ways to extract and enforce local specifications, including distillation, adaptive early stopping, or training-time internalization.

\paragraph{Prompting-based implementation.}
PSCI is training-free and immediately deployable, but prompting alone may be suboptimal for enforcing specifications reliably. Our residual-error analysis suggests that many remaining failures occur when the contract covers an obligation but the final answer does not execute it. This points to future work on stronger enforcement mechanisms, iterative repair, or training-time objectives that internalize specification compliance while reducing inference cost.

\section*{Ethics Statement}

PSCI is a test-time prompting scaffold evaluated on CL-Bench, a public benchmark whose contexts are expert-crafted through fictional creation or modification of existing content. No private user data is collected or generated at any stage. The specification contracts induced by PSCI are task-specific intermediate objects used only within the immediate generation pipeline; they are not stored, shared, or reused beyond evaluation. All evaluation uses the publicly released CL-Bench evaluation code and rubrics. We emphasize that improved context learning performance should not be interpreted as a guarantee of factual correctness or safety compliance in deployment, since specification acquisition addresses local validity within a provided context rather than global truthfulness.

Because PSCI extracts and enforces local specifications, it may also over-enforce brittle, biased, unsafe, or undesirable local rules if such rules appear in the context. This is especially important in legal, medical, financial, or institutional settings, where local policies may conflict with fairness, safety, or broader legal or ethical constraints. PSCI should therefore be used with expert oversight in high-stakes domains and should not be treated as a guarantee that the extracted specification is normatively appropriate.

\paragraph{Artifact licenses and intended use.}
CL-Bench is released under a custom evaluation-only license permitting use, modification, and distribution solely for model evaluation, testing, and benchmarking. Our use is fully consistent with this restriction: PSCI is a test-time prompting scaffold, and no stage of our pipeline trains, fine-tunes, distills, or otherwise updates model parameters on CL-Bench data. Evaluated models are accessed through official APIs or public weights and used for research benchmarking consistent with their respective terms.

\clearpage
\bibliography{iclr2026_conference}

@misc{dou2026clbench,
  title         = {{CL-bench}: A Benchmark for Context Learning},
  author        = {Dou, Shihan and Zhang, Ming and Yin, Zhangyue and Huang, Chenhao and Shen, Yujiong and Wang, Junzhe and Chen, Jiayi and Ni, Yuchen and Ye, Junjie and Zhang, Cheng and Xie, Huaibing and Hu, Jianglu and Wang, Shaolei and Wang, Weichao and Xiao, Yanling and Liu, Yiting and Xu, Zenan and Guo, Zhen and Zhou, Pluto and Gui, Tao and Wu, Zuxuan and Qiu, Xipeng and Zhang, Qi and Huang, Xuanjing and Jiang, Yu-Gang and Wang, Di and Yao, Shunyu},
  year          = {2026},
  eprint        = {2602.03587},
  archivePrefix = {arXiv},
  primaryClass  = {cs.CL}
}

@article{liu2024lost,
  title={Lost in the middle: How language models use long contexts},
  author={Liu, Nelson F and Lin, Kevin and Hewitt, John and Paranjape, Ashwin and Bevilacqua, Michele and Petroni, Fabio and Liang, Percy},
  journal={Transactions of the association for computational linguistics},
  volume={12},
  pages={157--173},
  year={2024}
}

@article{hsieh2024ruler,
  title={RULER: What's the real context size of your long-context language models?},
  author={Hsieh, Cheng-Ping and Sun, Simeng and Kriman, Samuel and Acharya, Shantanu and Rekesh, Dima and Jia, Fei and Zhang, Yang and Ginsburg, Boris},
  journal={arXiv preprint arXiv:2404.06654},
  year={2024}
}

@misc{kamradt2023niah,
  author = {Kamradt, Greg},
  title  = {Needle in a Haystack: Pressure Testing {LLMs}},
  year   = {2023},
  note   = {GitHub repository}
}

@inproceedings{bai2023longbench,
  title={Longbench: A bilingual, multitask benchmark for long context understanding},
  author={Bai, Yushi and Lv, Xin and Zhang, Jiajie and Lyu, Hongchang and Tang, Jiankai and Huang, Zhidian and Du, Zhengxiao and Liu, Xiao and Zeng, Aohan and Hou, Lei and others},
  booktitle={Proceedings of the 62nd annual meeting of the association for computational linguistics (volume 1: Long papers)},
  pages={3119--3137},
  year={2024}
}

@article{yan2024corrective,
  author  = {Shi-Qi Yan and Jia-Chen Gu and Yun Zhu and Zhen-Hua Ling},
  title   = {Corrective Retrieval Augmented Generation},
  journal = {arXiv preprint arXiv:2401.15884},
  year    = {2024}
}

@inproceedings{sarthi2024raptor,
  title={{RAPTOR}: Recursive Abstractive Processing for Tree-Organized Retrieval},
  author={Sarthi, Parth and Abdullah, Salman and Tuli, Aditi and Khanna, Shubh and Goldie, Anna and Manning, Christopher D.},
  booktitle={The Twelfth International Conference on Learning Representations (ICLR)},
  year={2024}
}

@article{wan2026inference,
  author  = {Yuxuan Wan and Tianqing Fang and Zaitang Li and Yintong Huo and Wenxuan Wang and Haitao Mi and Dong Yu and Michael R. Lyu},
  title   = {Inference-Time Scaling of Verification: Self-Evolving Deep Research Agents via Test-Time Rubric-Guided Verification},
  journal = {arXiv preprint arXiv:2601.15808},
  year    = {2026}
}

@inproceedings{raghavendra2026agentic,
  title={Agentic Rubrics as Contextual Verifiers for {SWE} Agents},
  author={Raghavendra, Mohit and Gunjal, Anisha and Liu, Bing and He, Yunzhong},
  booktitle={Proceedings of the 64th Annual Meeting of the Association for Computational Linguistics (Volume 1: Long Papers)},
  pages={15265--15290},
  year={2026}
}

@article{yue2025inference,
  title={Inference Scaling for Long-Context Retrieval Augmented Generation},
  author={Yue, Zhenrui and Zhuang, Honglei and Bai, Aijun and Hui, Kai and Jagerman, Rolf and Zeng, Hansi and Qin, Zhen and Wang, Dong and Wang, Xuanhui and Bendersky, Michael},
  journal={arXiv preprint arXiv:2410.04343},
  year={2024}
}

@article{zhou2023ifeval,
  title={Instruction-Following Evaluation for Large Language Models},
  author={Zhou, Jeffrey and Lu, Tianjian and Mishra, Swaroop and Brahma, Siddhartha and Basu, Sujoy and Luan, Yi and Zhou, Denny and Hou, Le},
  journal={arXiv preprint arXiv:2311.07911},
  year={2023}
}

@misc{xia2024fofo,
      title={FOFO: A Benchmark to Evaluate LLMs' Format-Following Capability}, 
      author={Congying Xia and Chen Xing and Jiangshu Du and Xinyi Yang and Yihao Feng and Ran Xu and Wenpeng Yin and Caiming Xiong},
      year={2024},
      eprint={2402.18667},
      archivePrefix={arXiv},
      primaryClass={cs.CL},
      url={https://arxiv.org/abs/2402.18667}, 
}

@misc{GeminiTeam2025,
  author       = {{Gemini Team}},
  title        = {Gemini 3 Pro Model Card},
  year         = {2025},
  howpublished = {\url{https://deepmind.google/models/model-cards/gemini-3-pro/}},
  note         = {Accessed: 2026-05-25}
}

@article{singh2025gpt5,
  title={{OpenAI} {GPT-5} System Card},
  author={Singh, Aaditya and Fry, Adam and Perelman, Adam and others},
  journal={arXiv preprint arXiv:2601.03267},
  year={2025}
}

@article{yang2025qwen3,
  title={Qwen3 technical report},
  author={Yang, An and Li, Anfeng and Yang, Baosong and Zhang, Beichen and Hui, Binyuan and Zheng, Bo and Yu, Bowen and Gao, Chang and Huang, Chengen and Lv, Chenxu and others},
  journal={arXiv preprint arXiv:2505.09388},
  year={2025}
}

@article{chen2025guideline,
  author  = {Jiaxiang Chen and Zhuo Wang and Mingxi Zou and Qifan Wang and Zenglin Xu},
  title   = {Guideline Forest: Retrieval-Augmented Reasoning with Branching Experience-Induced Guidelines},
  journal = {arXiv preprint arXiv:2506.07820},
  year    = {2025}
}

@inproceedings{asai2024selfrag,
  title={Self-{RAG}: Learning to Retrieve, Generate, and Critique through Self-Reflection},
  author={Asai, Akari and Wu, Zeqiu and Wang, Yizhong and Sil, Avirup and Hajishirzi, Hannaneh},
  booktitle={The Twelfth International Conference on Learning Representations (ICLR)},
  year={2024}
}

@article{edge2024graphrag,
  title={From local to global: A graph rag approach to query-focused summarization},
  author={Edge, Darren and Trinh, Ha and Cheng, Newman and Bradley, Joshua and Chao, Alex and Mody, Apurva and Truitt, Steven and Metropolitansky, Dasha and Ness, Robert Osazuwa and Larson, Jonathan},
  journal={arXiv preprint arXiv:2404.16130},
  year={2024}
}

@inproceedings{sun2023rankgpt,
  title={Is ChatGPT good at search? investigating large language models as re-ranking agents},
  author={Sun, Weiwei and Yan, Lingyong and Ma, Xinyu and Wang, Shuaiqiang and Ren, Pengjie and Chen, Zhumin and Yin, Dawei and Ren, Zhaochun},
  booktitle={Proceedings of the 2023 conference on empirical methods in natural language processing},
  pages={14918--14937},
  year={2023}
}

@article{shinn2023reflexion,
  title={Reflexion: Language agents with verbal reinforcement learning},
  author={Shinn, Noah and Cassano, Federico and Gopinath, Ashwin and Narasimhan, Karthik and Yao, Shunyu},
  journal={Advances in neural information processing systems},
  volume={36},
  pages={8634--8652},
  year={2023}
}

@article{madaan2023selfrefine,
  title={Self-refine: Iterative refinement with self-feedback},
  author={Madaan, Aman and Tandon, Niket and Gupta, Prakhar and Hallinan, Skyler and Gao, Luyu and Wiegreffe, Sarah and Alon, Uri and Dziri, Nouha and Prabhumoye, Shrimai and Yang, Yiming and others},
  journal={Advances in neural information processing systems},
  volume={36},
  pages={46534--46594},
  year={2023}
}

@article{cook2024tick,
  title={Ticking all the boxes: Generated checklists improve llm evaluation and generation},
  author={Cook, Jonathan and Rockt{\"a}schel, Tim and Foerster, Jakob and Aumiller, Dennis and Wang, Alex},
  journal={arXiv preprint arXiv:2410.03608},
  year={2024}
}

@article{wei2022cot,
  title={Chain-of-thought prompting elicits reasoning in large language models},
  author={Wei, Jason and Wang, Xuezhi and Schuurmans, Dale and Bosma, Maarten and Xia, Fei and Chi, Ed and Le, Quoc V and Zhou, Denny and others},
  journal={Advances in neural information processing systems},
  volume={35},
  pages={24824--24837},
  year={2022}
}

@article{packer2023memgpt,
  author  = {Packer, Charles and others},
  title   = {{MemGPT}: Towards {LLMs} as Operating Systems},
  journal = {arXiv preprint arXiv:2310.08560},
  year    = {2023}
}

@article{kojima2022zeroshot,
  title={Large language models are zero-shot reasoners},
  author={Kojima, Takeshi and Gu, Shixiang Shane and Reid, Machel and Matsuo, Yutaka and Iwasawa, Yusuke},
  journal={Advances in neural information processing systems},
  volume={35},
  pages={22199--22213},
  year={2022}
}

@article{khattab2024dspy,
  title={Dspy: Compiling declarative language model calls into self-improving pipelines},
  author={Khattab, Omar and Singhvi, Arnav and Maheshwari, Paridhi and Zhang, Zhiyuan and Santhanam, Keshav and Vardhamanan, Sri and Haq, Saiful and Sharma, Ashutosh and Joshi, Thomas T and Moazam, Hanna and others},
  journal={arXiv preprint arXiv:2310.03714},
  year={2023}
}

@article{saunders2022selfcritique,
  title={Self-critiquing models for assisting human evaluators},
  author={Saunders, William and Yeh, Catherine and Wu, Jeff and Bills, Steven and Ouyang, Long and Ward, Jonathan and Leike, Jan},
  journal={arXiv preprint arXiv:2206.05802},
  year={2022}
}

@article{lewis2020rag,
  title={Retrieval-augmented generation for knowledge-intensive nlp tasks},
  author={Lewis, Patrick and Perez, Ethan and Piktus, Aleksandra and Petroni, Fabio and Karpukhin, Vladimir and Goyal, Naman and K{\"u}ttler, Heinrich and Lewis, Mike and Yih, Wen-tau and Rockt{\"a}schel, Tim and others},
  journal={Advances in neural information processing systems},
  volume={33},
  pages={9459--9474},
  year={2020}
}

@article{gao2023retrieval,
  title={Retrieval-augmented generation for large language models: A survey},
  author={Gao, Yunfan and Xiong, Yun and Gao, Xinyu and Jia, Kangxiang and Pan, Jinliu and Bi, Yuxi and Dai, Yixin and Sun, Jiawei and Wang, Haofen and Wang, Haofen and others},
  journal={arXiv preprint arXiv:2312.10997},
  volume={2},
  number={1},
  pages={32},
  year={2023}
}

@article{wang2023selfconsistency,
  title={Self-consistency improves chain of thought reasoning in language models},
  author={Wang, Xuezhi and Wei, Jason and Schuurmans, Dale and Le, Quoc and Chi, Ed and Narang, Sharan and Chowdhery, Aakanksha and Zhou, Denny},
  journal={arXiv preprint arXiv:2203.11171},
  year={2022}
}

@article{yao2023react,
  title={React: Synergizing reasoning and acting in language models},
  author={Yao, Shunyu and Zhao, Jeffrey and Yu, Dian and Du, Nan and Shafran, Izhak and Narasimhan, Karthik and Cao, Yuan},
  journal={arXiv preprint arXiv:2210.03629},
  year={2022}
}
\bibliographystyle{iclr2026_conference}

\appendix

\section{Supplementary Material for Empirical Study}
\label{app:study}

\subsection{CL-Bench Design and Contamination Prevention}
\label{app:clbench}

CL-Bench~\citep{dou2026clbench} prevents contamination through three approaches: (1)~\emph{fictional creation}---experts construct entirely novel content such as complete legal systems for fictional countries or new programming languages with unique syntax; (2)~\emph{modification of existing content}---altering real-world knowledge such as changing scientific definitions, historical events, or technical specifications; and (3)~\emph{incorporation of niche and emerging content}---using cutting-edge or narrow-domain knowledge not well-represented in pre-training corpora. Each context requires approximately 20 hours of expert annotation effort and undergoes multiple rounds of quality review.

A context-free ablation confirms the effectiveness of this design: GPT-5.1 solves less than 1\% of tasks when the context is removed~\citep{dou2026clbench}, confirming that CL-Bench tasks cannot be solved from pre-trained knowledge alone.

\subsection{Classification Protocol Details}
\label{app:protocol}

The LLM-based classifier (GPT-5.1, the same model used by CL-Bench for rubric evaluation) receives the full task context and rubric text and labels each item as content or specification. The deterministic rule-based classifier combines lexical cues (e.g., ``must include,'' ``format as''), structural cues (e.g., schema references, enumerated constraints), and semantic cues (e.g., whether the item targets factual derivation or behavioral compliance). It includes an explicit abstention label for items where no rule fires with sufficient confidence.

Under the agreement protocol, an item is labeled specification or content only when both classifiers independently produce the same label. Items where the classifiers disagree or where the deterministic classifier abstains are labeled \emph{ambiguous} and excluded from both counts. This conservative design ensures that the reported 55.4\% specification share and 22.6\% content share represent high-precision estimates rather than aggressive forced-choice labels.

To validate the taxonomy, three PhD-level annotators independently audit a random sample of 300 rubric items. Annotators are shown the task context and rubric item, but not the automatic classifier labels. They assign each item to content, specification, or ambiguous using the same operational definitions. Final human labels are determined by majority vote. On items assigned non-ambiguous labels by the conservative automatic protocol, the human majority label agrees with the automatic label in over 90\% of cases, supporting the reliability of the taxonomy.

\subsection{Near-Miss Validation}
\label{app:nearmiss}

To confirm that the 281 single-item-failure tasks are genuine near-misses, we remove the single failed rubric item from each task and re-run the full generation and judging pipeline. After three independent runs, 99\% of these tasks pass, confirming that the identified rubric item is indeed the sole point of failure.

This finding is further consistent with CL-Bench's own error analysis, which reports format errors at 35\%+ and context misuse exceeding 60\% of failures even for top-performing models~\citep{dou2026clbench}.

\subsection{Traceability Annotation Protocol}
\label{app:traceability}

For each specification item, we annotate whether the obligation can be traced to the task or context, and through which evidentiary channel:

\begin{itemize}
    \item \textbf{Explicit declaration} (71.6\%): The context directly states the requirement (e.g., ``if a gap is detected, emit \texttt{Sequence\_Gap\_Warning}'').
    \item \textbf{Validation pattern} (21.1\%): The context demonstrates a validation procedure or schema implying what the output must satisfy.
    \item \textbf{Implicit example} (2.0\%): The context includes a worked example whose format implies the expected behavior.
    \item \textbf{Repeated convention} (0.8\%): The context uses a consistent structure repeatedly, implying the answer should follow the same pattern.
    \item \textbf{Not derivable} (0.4\%): The obligation cannot be traced to the task or context.
    \item \textbf{Too short to judge} (4.1\%): The rubric snippet is too brief to determine traceability.
\end{itemize}

To validate these annotations, three PhD-level annotators each independently classified 100 randomly sampled specification items, with final labels determined by majority vote. Human annotators confirmed traceability at rates consistent with or higher than the automated system, supporting the reliability of the reported 95.5\% rate.

\subsection{Content vs.\ Specification: Worked Examples}
\label{app:examples}

\paragraph{Content item.} ``The response should identify Event ID 4072 as the anomalous entry.'' The verdict depends on whether the answer states the correct proposition. Any format communicating this fact satisfies the item.

\paragraph{Specification item.} ``When a reserved event ID is skipped, the log must emit the flag \texttt{Sequence\_Gap\_Warning}.'' The verdict depends on conforming to a locally defined logging convention. Identifying the gap but logging it as ``missing\_id'' would fail despite containing the correct content.

\paragraph{Ambiguous item.} ``The response should include a summary table of all detected anomalies with timestamps.'' This evaluates both content (which anomalies) and specification (table format). Under our protocol, such items are labeled ambiguous.

\section{Supplementary Material for Method}
\label{app:method}

\subsection{Rotated Exemplar Design}
\label{app:exemplars}

The contract induction step uses a small set of rotated demonstrations to teach the style and granularity of specification contracts. For each evaluation task, demonstrations are drawn from other disjoint context learning tasks and never from the current task. Thus, the current task's answer, public rubric, context, or task-specific contract is never shown during induction. The final contract used for generation is induced from the current task context. Each demonstration illustrates how to express local obligations---such as required fields, validation checks, exact artifacts, exclusions, ordering constraints, and completeness rules---but does not provide task-specific information for the current example. We use 10 demonstrations per task. Zero-shot induction without demonstrations still improves over the baseline (A8/A11 in Table~\ref{tab:ablation}), while generic and shuffled specifications underperform, showing that demonstrations refine contract granularity but do not create the effect.

\subsection{Design Decisions: Extended Discussion}
\label{app:design}

Two design decisions distinguish PSCI from generic self-refinement and merit explicit justification.

\paragraph{Contract before answer.} The contract $K$ is induced before the draft $y_0$, so it functions as a control object that shapes generation rather than a post-hoc grading rubric applied to a finished draft. This is motivated by the traceability analysis in \autoref{sec:traceability}: since 76.7\% of specifications are absent from the query, they must be surfaced before generation begins. If the model generates first, it anchors to a query-driven structure that post-hoc repair cannot fully restructure. Ablation A6 confirms this empirically: answer-first ordering collapses to 23.67\% strict pass and 70.67\% rubric, which remains far below PSCI, and its rubric pass falls below the direct baseline.

\paragraph{Checker targets the contract.} The checker assesses the draft against the induced contract $K$, not against open-ended quality criteria. This design prevents the checker from simply restating what the answer already says and ensures that verification targets specific obligations. Ablation A3 (checker + repair without contract, 24.32\%) versus PSCI (28.14\%) confirms that grounding verification in a specification contract is essential.

\subsection{Prompt Templates}
Please see \autoref{app:prompts} for the full prompt that can reproduce PSCI.

\subsection{Evaluation Protocol Details}
\label{app:eval}

\paragraph{Verifier.}
We use the exact evaluation code and GPT-5.1-based verifier released by the CL-Bench authors~\citep{dou2026clbench}. Each rubric item is evaluated as a binary yes/no question; a task is passed only when all associated items receive ``yes.'' The CL-Bench authors report over 90\% agreement between GPT-5.1 judgments and human annotations in their validation study, and further show that raw agreement between GPT-5.1 and two alternative verifiers (Claude Opus 4.5, Qwen-3-Max) exceeds 90\%, indicating minimal self-evaluation bias.

\paragraph{Run protocol.}
For each method and model combination, we conduct three independent runs on the full 1{,}899-task set and report the mean along with standard deviation. For PSCI, the complete end-to-end pipeline ($\mathrm{Induce} \to \mathrm{Generate} \to \mathrm{Check} \to \mathrm{Repair}$) constitutes a single run, reflecting real-world deployability where the four stages execute sequentially without human intervention. We apply reruns only for API infrastructure failures (rate-limit errors, timeouts, connection drops); no selective filtering, output inspection, or cherry-picking is performed at any stage.

\paragraph{Baseline implementation.}
All twelve baselines are implemented following the original papers as closely as possible and evaluated under the same GPT-5.1 medium-thinking setting. For methods requiring retrieval (Self-RAG, CRAG, RAPTOR, GraphRAG, RankGPT, IterDRAG), the retrieval corpus is the task's own context---no external knowledge base is used, consistent with CL-Bench's design that all required knowledge resides in the provided context. SimpleMem uses a memory-style compression of the context. Full implementation details for each baseline are provided in Appendix~\ref{app:baselines}.

\paragraph{Compute budget.}
  PSCI is an inference-time scaling method and is therefore more expensive than single-pass generation: it uses four sequential model calls per task for
  specification induction, contract-guided answer generation, contract-aware checking, and targeted repair. We treat this as a central tradeoff rather than a
  hidden cost. Table~\ref{tab:compute_budget} reports both latency and token usage. To test whether PSCI's gains are merely a consequence of additional
  computation, we include compute-matched and compute-heavier controls, including generic self-reflection, iterative retrieval, guideline-based reasoning, and
  verification-style baselines. These controls use comparable or larger inference budgets but do not match PSCI, indicating that the gains come from how
  computation is structured---around task-specific specification induction and enforcement---rather than from extra tokens alone.

\begin{table*}[t]
  \centering
  \small
  \begin{tabular}{cccccc}
  \toprule
  Method & LLM stages/task & Gen. tokens/task & Latency & Strict pass & Rubric pass \\
  \midrule
  GPT Baseline & 1 & 18.9k & 44.6s & 22.55 & 76.01 \\
  Baseline + SI & 2 & 56.1k & 54.9s & 26.09 & 79.10 \\
  PSCI & 4 & 91.6k & 132.2s & 28.14 & 81.13 \\
  \midrule
  Compute-Matched Majority Vote & 5 & 94.5k & 223s & 24.12 & 74.23 \\
  Compute-Matched Self-Reflection & 5 & 94.5k & 254s & 23.02 & 71.24 \\
  \midrule
  Reflexion & 3 & 47.2k & 108.5s & 22.33 & 71.64 \\
  IterDRAG & multiple/iterative & 70.8k & 166.6s & 22.00 & 72.10 \\
  Guideline Forest & multiple/branching & 84.5k & 246.6s & 22.00 & 73.82 \\
  TICK/STICK & multiple/variable & 67.9k & 165.5s & 24.00 & 74.20 \\
  DeepVerifier & multiple/variable & 78.8k & 155.4s & 21.33 & 70.55 \\
  Agentic Rubrics & multiple/variable & 84.8k & 260.2s & 20.33 & 69.99 \\
  \bottomrule
  \end{tabular}
\caption{
Compute and performance comparison. Token counts exclude judge calls and report generated tokens across method stages. Latency is wall-clock time per task under the same API/model setting. PSCI is more expensive than single-pass generation, but compute-matched or compute-heavier alternatives do not recover its gains, indicating that performance comes from task-specific specification induction and enforcement rather than inference budget alone.
}
  \label{tab:compute_budget}
\end{table*}

\section{Additional Baseline Details}
\label{app:baselines}

We implement each baseline as an inference-time prompting or retrieval procedure over the same CL-Bench instance. Unless otherwise noted, all methods receive the task instruction and may access only the provided task context. No method is given the public evaluation rubrics.

\paragraph{Self-RAG.}
Self-RAG~\citep{asai2024selfrag} trains or prompts a model to decide when retrieval is needed and to use retrieved evidence when generating an answer. In our implementation, we split the task context into retrievable chunks, prompt the model to identify evidence needs, retrieve relevant chunks from the local context, and condition the final response on that evidence.

\paragraph{CRAG.}
Corrective RAG~\citep{yan2024corrective} improves retrieval-augmented generation by adding a correction mechanism that evaluates whether retrieved documents are relevant, incomplete, or misleading. We adapt this idea by retrieving candidate context chunks, asking the model to judge evidence sufficiency, and performing an additional corrective retrieval pass before final answer generation when needed.

\paragraph{RankGPT.}
RankGPT~\citep{sun2023rankgpt} uses an LLM as a listwise reranker, relying on the model's semantic judgment to reorder candidate documents by relevance. We first retrieve candidate chunks from the task context using lexical similarity, then prompt GPT-5.1 to rank them by usefulness for the task, and finally generate the answer from the top-ranked evidence.

\paragraph{RAPTOR.}
RAPTOR~\citep{sarthi2024raptor} builds a recursive tree of summaries so that long contexts can be queried at multiple levels of abstraction. We chunk each CL-Bench context, summarize groups of chunks into higher-level nodes, retrieve both fine-grained chunks and abstract summaries, and generate the final answer from the resulting evidence set.

\paragraph{GraphRAG.}
GraphRAG~\citep{edge2024graphrag} represents long documents as a graph of entities, relations, and communities, enabling retrieval over both source text and structured relations. We construct a lightweight graph from each task context by extracting entities, events, constraints, and links, retrieve relevant graph neighborhoods with associated source snippets, and use them to produce the answer.

\paragraph{TICK/STICK.}
TICK/STICK~\citep{cook2024tick} is a structured reasoning baseline that decomposes a task into intermediate checks before producing the final answer. We implement it by prompting the model to extract task constraints, identify relevant context evidence, enumerate intermediate validation checks, and then synthesize a final response from this checklist-style decomposition.

\paragraph{DeepVerifier.}
DeepVerifier~\citep{wan2026inference} is a verification-oriented baseline that separates answer generation from answer checking. The model first drafts an answer, then performs a second pass that inspects the draft against the task and retrieved context, identifies omissions or contradictions, and produces a revised final answer.

\paragraph{Agentic Rubrics.}
Agentic Rubrics~\citep{raghavendra2026agentic} asks an agent to infer likely evaluation criteria before answering, approximating a rubric without seeing the true public rubric. We prompt the model to derive a private checklist from the task and context, then generate the answer using this checklist; unlike PSCI, it does not include the full contract-induction and contract-aware checker-repair pipeline.

\paragraph{SimpleMem.}
SimpleMem uses a compact memory representation to retain task-relevant information from long contexts. We ask the model to compress each CL-Bench context into a memory of salient facts, constraints, procedures, and state, then answer from the task plus this memory.%
\footnote{SimpleMem is our in-house implementation inspired by memory-augmented LLM approaches such as MemGPT~\citep{packer2023memgpt}.}

\paragraph{Reflexion.}
Reflexion~\citep{shinn2023reflexion} improves agents through verbal self-feedback, where the model critiques its own prior behavior and uses the critique to revise future actions. We implement a single-instance version: the model drafts an answer, writes a natural-language reflection about possible failures relative to the task and context, and then revises the answer.

\paragraph{IterDRAG.}
IterDRAG~\citep{yue2025inference} scales inference for retrieval-augmented generation by alternating reasoning with additional retrieval steps. We implement an iterative loop in which the model reasons about missing information, retrieves more local context chunks, updates its working evidence, and then produces a final answer.

\paragraph{Guideline Forest.}
Guideline Forest~\citep{chen2025guideline} uses branching experience-induced guidelines to guide reasoning through multiple candidate solution paths. We approximate this by inducing several reusable guidelines from the task context, applying them along parallel reasoning branches, and merging the resulting drafts into a final response.
  
\section{Additional result visualizations.}
\subsection{Visualization of ablation study results.} 

\begin{figure*}[t]
  \centering
  \includegraphics[width=1\linewidth]{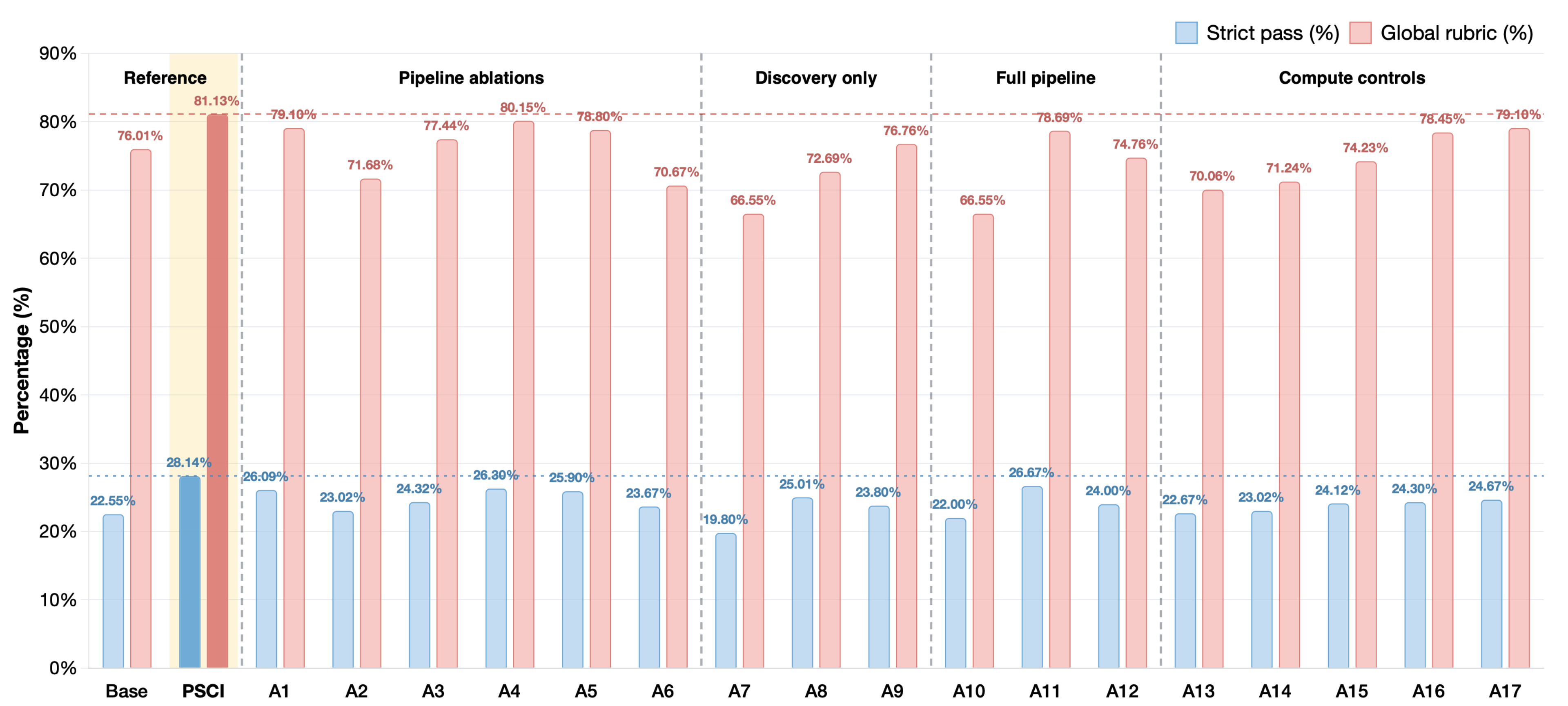}
\caption{Ablation results (strict pass and global rubric \%). Three key tests rule out alternative explanations: shuffled contracts collapse below baseline (A7/A10), answer-first ordering fails (A6), and spec-free critique returns to baseline (A13).}
  \label{fig:ablations}
\end{figure*}

\subsection{Human Validation of Contract--Rubric Overlap}
\label{app:human_overlap}

To validate the automated overlap measurement between induced contracts and public rubrics, three PhD-level annotators independently labeled the same 100 randomly sampled non-ambiguous rubric items. For each item, annotators judged whether the induced contract strongly covered, partially covered, contradicted, or did not cover the corresponding public rubric item. Final labels were determined by majority vote.

Human annotators assigned strong coverage to 64.95\% of items, compared with 60.8\% from the automated overlap system. Thus, the automated measurement used in the main paper is conservative. This supports the claim that PSCI recovers most of the public evaluation surface without ever observing the public rubric.

\section{Extended Related Work}
\label{app:related}

\subsection{Long-Context Evaluation, Instruction Following, and Context Learning}
\label{app:related_long_context}

Long-context evaluation has traditionally asked whether models can use information distributed across long inputs. Early evaluations focused on retrieval-style probes, including needle-in-a-haystack~\citep{kamradt2023niah} and controlled recall or aggregation tasks~\citep{hsieh2024ruler}. Broader benchmarks such as LongBench~\citep{bai2023longbench} and analyses of position sensitivity~\citep{liu2024lost} evaluate long-document understanding, question answering, summarization, and reasoning over extended contexts. These benchmarks are essential for measuring evidence use, but the task interface is usually fixed: the model must answer a question, summarize, retrieve, aggregate, or reason over provided evidence.

Context learning differs in that the context defines the local task system itself. A model must learn concepts, rules, procedures, empirical regularities, and answer validity conditions supplied at inference time. CL-Bench~\citep{dou2026clbench} formalizes this setting through tasks whose required knowledge is absent from pre-training and embedded in long, expert-written contexts. We focus on CL-Bench because it is currently the only large-scale benchmark explicitly designed for context learning, and because its strict public rubrics make local answer obligations measurable.

Specification acquisition is also related to instruction following and format adherence. Benchmarks such as IFEval~\citep{zhou2023ifeval} and format-following evaluations~\citep{xia2024fofo} test whether models obey explicit constraints. In contrast, CL-Bench specifications are often not stated in the user query. They are query-implicit, distributed across local documentation, and may appear as schemas, examples, validation patterns, edge-case rules, logging conventions, or procedural requirements. Therefore, the challenge is not simply following an explicit instruction, but inducing the local specification before answering.

\subsection{Retrieval, Memory, and Context Restructuring}
\label{app:related_retrieval}

A central line of work improves content access through retrieval-augmented generation~\citep{lewis2020rag,gao2023retrieval}, adaptive retrieval and correction~\citep{asai2024selfrag,yan2024corrective}, reranking~\citep{sun2023rankgpt}, memory systems~\citep{packer2023memgpt}, hierarchical context organization~\citep{sarthi2024raptor,edge2024graphrag}, and iterative retrieval~\citep{yue2025inference}. These methods address an important part of context learning: a model must find relevant evidence before it can reason over it.

However, our empirical results show that content access is insufficient for CL-Bench. Retrieval-oriented methods tend to surface query-relevant, answer-bearing evidence. In contrast, many local specifications are validity-relevant rather than answer-bearing: they govern what the final response must include, exclude, validate, log, order, ground, or complete. Such obligations are often low-salience and scattered across background conventions, schemas, exceptions, logging rules, or validation requirements. Thus, a method may retrieve the content needed to answer the apparent query while missing the local obligations required for an acceptable answer. This explains why the retrieval and context-restructuring baselines in \autoref{sec:main} do not reliably improve over full-context prompting.

\subsection{Critique, Verification, and Structured Intermediate Computation}
\label{app:related_verification}

Several methods improve outputs through critique, reflection, or revision. Self-Refine~\citep{madaan2023selfrefine}, Reflexion~\citep{shinn2023reflexion}, and self-critique~\citep{saunders2022selfcritique} ask models to identify and repair their own errors. Rubric-aware and checklist-based approaches, including TICK/STICK~\citep{cook2024tick}, DeepVerifier~\citep{wan2026inference}, and Agentic Rubrics~\citep{raghavendra2026agentic}, use generated or fixed criteria to verify outputs. These methods are close in spirit to PSCI, but they usually treat the checklist or critique as generic, fixed, or post-hoc.

PSCI instead treats the local specification as a latent variable induced from the task context before generation. The induced contract is not merely a verifier prompt; it is a task-specific set of obligations used as a shared control object for generation, checking, and repair. This distinction is empirically important: compute-matched generic critique returns to baseline, checker/repair without a specification contract underperforms, answer-first contract induction fails, and shuffled contracts collapse below baseline. Therefore, the gain does not come from extra calls or generic verification, but from inducing the right local specification and enforcing it at the right stage.

More broadly, PSCI belongs to a family of methods that introduce structured intermediate computation, such as chain-of-thought~\citep{wei2022cot,kojima2022zeroshot}, self-consistency~\citep{wang2023selfconsistency}, ReAct~\citep{yao2023react}, and declarative language-model pipelines~\citep{khattab2024dspy}. The difference is again the intermediate object: PSCI does not only elicit reasoning traces or tool actions; it induces a context-grounded obligation set that defines local answer validity. Our ablations show that this object, rather than structured prompting alone, is what drives improvement on CL-Bench.

\section{Human Audit of Baseline-to-PSCI Flips}
\label{app:human_flips}

To verify that PSCI gains are not an artifact of the automatic CL-Bench verifier, we conduct a blind human audit on tasks where the automatic evaluator marks the direct baseline as failed and PSCI as passed. We randomly sample 100 such baseline-fail/PSCI-pass tasks from the full evaluation set, covering all four CL-Bench task families when possible.

\paragraph{Annotators.}
Three PhD-level annotators independently evaluate each sampled task. Annotators are familiar with LLM evaluation and rubric-based assessment, but are not told which answer is produced by the baseline or by PSCI. Answer order is randomized independently for each task.

\paragraph{Materials.}
For each task, annotators are shown the task query, the task context, all public CL-Bench rubric items associated with that task, and two anonymized model answers. The automatic verifier decision, failed rubric items, and method identity are hidden. Annotators are instructed to judge correctness according to the provided rubric, not fluency, verbosity, or style unless explicitly required by the rubric.

\paragraph{Annotation questions.}
For each task, annotators answer two questions:
\begin{enumerate}
    \item \textbf{Preference.} Which answer better satisfies the task and rubric? Options are Answer A, Answer B, Tie, or Unable to judge.
    \item \textbf{Flip correctness.} Which automatic flip status best describes the two answers? Options are: Answer A passes while Answer B fails, Answer B passes while Answer A fails, both pass, both fail, or Unable to judge.
\end{enumerate}
After annotation, answer identities are mapped back to baseline and PSCI. A task is counted as confirming the automatic flip if the majority label indicates that the PSCI answer passes while the baseline answer fails. Final task-level labels are determined by majority vote across the three annotators.

\paragraph{Results.}
Human judgments strongly support the automatic improvement signal. By majority vote, annotators prefer the PSCI answer in 96/100 tasks (96\%) and confirm the automatic baseline-fail/PSCI-pass flip in 92/100 tasks (92\%). Thus, most automatic PSCI wins correspond to genuine task-level improvements under human evaluation rather than verifier artifacts.

\paragraph{Inter-annotator agreement.}
For answer preference, all three annotators agree on 92/100 tasks. For flip correctness, all three annotators agree on 90/100 tasks. The remaining cases are resolved by majority vote. We exclude ``Unable to judge'' responses from agreement computation and observe no systematic disagreement concentrated in any single CL-Bench task family.

\paragraph{Interpretation.}
This audit is targeted rather than exhaustive: it focuses on the most important possible evaluator artifact, namely whether automatic baseline-fail/PSCI-pass flips reflect real improvements. The high PSCI preference rate and high flip-confirmation rate indicate that the reported gains are not merely due to PSCI producing more rubric-like text for the automatic verifier. Humans also judge PSCI outputs as more correct on the sampled task flips.

\section{Additional material for discussion}

\subsection{Coverage Label Definitions and Examples}
\label{app:coverage}

Each public rubric item is compared against the privately induced contract and assigned one of four coverage labels. We illustrate with the \texttt{Sequence\_Gap\_Warning} example from \autoref{sec:intro}.

\begin{itemize}
    \item \textbf{Strongly covered.} The contract entails the same condition and required action as the rubric item. \emph{Example:} Contract states ``if a reserved event ID is skipped, emit \texttt{Sequence\_Gap\_Warning}''; rubric requires the same.
    \item \textbf{Partially covered.} The contract captures the main obligation but misses scope, exactness, or an edge condition. \emph{Example:} Contract states ``log all sequence gaps'' but does not specify the exact flag \texttt{Sequence\_Gap\_Warning}.
    \item \textbf{Contradicted.} The contract imposes an incompatible obligation. \emph{Example:} Contract states ``no warning is needed for reserved-ID skips''; rubric requires the warning.
    \item \textbf{Uncovered.} No contract item addresses the obligation. \emph{Example:} No contract item mentions sequence gaps or reserved-ID handling.
\end{itemize}

Coverage labels are assigned automatically by prompting GPT-5.1 with the rubric item and corresponding contract, then validated against the human annotations described in \autoref{app:human_overlap}.

\subsection{Extended Discussion: Why Contracts Must Precede Generation}
\label{app:control_object}

The answer-first ablation isolates whether the same specification-induction machinery can be used only after an initial answer has already been produced. This setting underperforms because the contract no longer controls the construction of the answer. In context learning, most local specifications are not directly inferable from the user query; they are distributed across background schemas, examples, validation rules, exceptions, and procedural conventions. Therefore, a direct answer is often organized around the apparent query objective before these obligations become explicit constraints.

Once this query-driven answer structure is fixed, repair is limited. It can add a missing label, patch a local field, or remove an unsupported claim, but it is less reliable at changing the answer's global organization: which fields are present, which evidence is cited, which cases are enumerated, and what completeness means. This explains why answer-first ordering and self-generated rubric verification underperform PSCI. They apply specification-like checks after generation, when many validity-relevant decisions have already been made.

In contrast, PSCI induces the contract before answering, so the specification acts as a control object shared by generation, checking, and repair. This does not mean post-hoc verification is useless; rather, verification is most effective when it checks compliance with obligations that have already shaped the draft. Thus, the main benefit of PSCI is not merely producing a better grader, but making local specifications active before the answer is formed.

\section{Reproducibility: PSCI Prompt Templates}
\label{app:prompts}

We provide the prompt templates used for PSCI. The public rubric for the current evaluation task is never shown to any stage. Curly-braced fields denote instance-specific content. For the few-shot induction variant, we use rotated demonstrations only to teach the style and granularity of specification contracts. For each evaluation task, demonstrations are drawn from other context learning tasks; no same-task answer, public rubric, or task-specific contract is shown. Demonstrations are therefore used as formatting/style calibration, not as a source of task-specific specifications.

\paragraph{Contract induction.}
PSCI induces a private specification contract before answering. We run a source-only contract prompt and a rotated-demonstration contract prompt, then deterministically merge and de-duplicate their contract items.

\noindent\textbf{Source-only contract prompt.}
\begin{lstlisting}[language={},breaklines=true,breakatwhitespace=false,basicstyle=\ttfamily\scriptsize]
System:
You are a context learning specification analyst. Your job is to infer the local answer specification from the source context and task. Do not answer the task. Do not use hidden benchmark rubrics or gold answers. Return strict JSON only.

User:
Infer a private specification contract for a correct answer to the task. The contract should capture what a locally valid answer must satisfy: required fields, validation steps, edge cases, source-specific schemas, sequencing, exact-format obligations, evidence expectations, required actions, exclusions, and completeness rules.

Important constraints:
- Derive obligations from the provided context and task, not from generic prior habits.
- Do not provide the final answer.
- Avoid leaking task-specific final values, dates, names, winners, classifications, computed results, or conclusions unless they are fixed format/schema constants explicitly named by the context.
- It is allowed to name exact required local artifacts, labels, fields, or warning codes if the context defines them as output obligations.
- Prefer source-relative wording such as "the source-specified status field" or "all context-defined validation checks".
- Include obligations that are likely to be missed by direct answering.
- Keep the contract concise enough to be useful as an answer checklist.

Return JSON exactly as:
{
  "contract_items": [
    {"id": "C1", "criterion": "...", "source_basis": "context/task signal that implies this standard"},
    {"id": "C2", "criterion": "...", "source_basis": "..."}
  ],
  "answer_shape": "brief description of expected output shape",
  "risk_notes": ["likely omission or edge case"]
}

### Source Context
{source_context}

### Task
{task}
\end{lstlisting}

\noindent\textbf{Rotated-demonstration contract prompt.}
\begin{lstlisting}[language={},breaklines=true,breakatwhitespace=false,basicstyle=\ttfamily\scriptsize]
System:
You are a context learning specification analyst. Infer the local answer specification from the source context and task. Do not answer the task. You may study the provided rubric-style demonstrations from other context learning tasks to learn the style and granularity of specification standards, but you must not use or assume any hidden rubric for the current task. Return strict JSON only.

User:
You will see rubric-style demonstrations from other context learning tasks. Use them only as demonstrations of what strict local specifications look like: explicit format constraints, local field names, edge cases, required validations, exact strings, source-grounding obligations, exclusions, and completeness criteria. Do not copy their domain content unless it is also present in the current source context.

For the current task, infer a private specification contract from the current source context and current task only. The contract should be non-answer-revealing: it may name required local fields, formats, exact labels, and validation procedures, but it should not disclose final computed values, case outcomes, winners, classifications, or other final answers unless those are fixed constants explicitly named by the context.

Return JSON exactly as:
{
  "contract_items": [
    {"id": "C1", "criterion": "...", "source_basis": "context/task signal that implies this standard"},
    {"id": "C2", "criterion": "...", "source_basis": "..."}
  ],
  "answer_shape": "brief description of expected output shape",
  "risk_notes": ["likely omission or edge case"]
}

Write 10-20 precise, enforceable contract items for complex tasks. Prefer items that a strict grader could check directly. Avoid generic advice.

### Rotated rubric-style demonstrations from other context learning tasks
{heldout_rotated_rubric_demonstrations}

### Current Source Context
{source_context}

### Current Task
{task}
\end{lstlisting}

\paragraph{Answer generation.}
The answerer receives the original task, source context, and induced private contract. The contract guides generation, but the source context and task remain authoritative.

\begin{lstlisting}[language={},breaklines=true,breakatwhitespace=false,basicstyle=\ttfamily\scriptsize]
System:
You are a source-grounded context learning answerer. You receive an inferred private specification contract, but it may contain mistakes. The source context and original task are authoritative. Silently audit the contract against the context/task, use only grounded obligations, recover any obvious local specifications the contract missed, then return only the final answer.

User:
Produce the final answer to the original task.

Silent audit rules before writing:
1. Verify every private-contract item against the source context/task. Follow grounded items.
2. If a contract item contradicts a clearer source-context or task rule, obey the source/task.
3. Pay special attention to exact strings, named sections, ordering, numbering, persona/system rules, schemas, units, citations, refusal phrases, source-only requirements, exclusions, and completeness conditions.
4. If the task says to use provided documents, avoid unsupported domain additions and unnecessary advanced details.
5. If drafts are supplied, use them only as candidate wording/evidence. Do not preserve their errors.
6. Do not mention the private contract, rubric, audit, or this process.

### Source Context
{source_context}

### Original Task
{task}

### Inferred Private Specification Contract
{private_contract}

### Draft Answers, If Any
{draft_answers_or_empty_list}

### Output
Return the final answer only.
\end{lstlisting}

\paragraph{Adversarial checking.}
The checker inspects the draft answer against the private specification contract using visible evidence only. It does not see the public rubric or gold answer.

\begin{lstlisting}[language={},breaklines=true,breakatwhitespace=false,basicstyle=\ttfamily\scriptsize]
System:
You are a strict private-contract checker for long-context context learning tasks. You do not know public rubrics or gold answers. Return strict JSON only.

User:
Perform an adversarial item-by-item visibility check.

For every private contract item, construct the concrete visible test that the final answer must pass. Then mark it pass/fail/uncertain. A pass requires an exact quote or precise paraphrase from the answer; do not give credit for intent, implication, or facts present only in the source context. If the answer is missing the required field/string/action/calculation/caveat, mark fail. If the answer may satisfy it but you cannot point to answer evidence, mark uncertain.

Also list context-obvious obligations that appear missing or weakly represented in the private contract. These must cite a task/context basis.

Return JSON exactly as:
{
  "overall_assessment": "brief",
  "item_checks": [
    {
      "private_index": 1,
      "test": "visible requirement",
      "status": "pass|fail|uncertain",
      "answer_evidence": "short quote or NONE",
      "source_basis": "task/context/private basis",
      "repair": "minimal edit if fail/uncertain",
      "priority": "high|medium|low"
    }
  ],
  "latent_gap_tests": [
    {
      "test": "required by task/context",
      "answer_evidence": "quote or NONE",
      "source_basis": "task/context quote",
      "repair": "minimal edit",
      "priority": "high|medium|low"
    }
  ],
  "source_only_risks": [
    {
      "claim": "answer claim possibly unsupported",
      "risk": "why risky",
      "repair": "remove/qualify/ground",
      "priority": "high|medium|low"
    }
  ],
  "repair_plan": ["up to {max_repairs} concrete edits, ordered by expected strict-rubric impact"],
  "preserve": ["answer parts that should stay unchanged"]
}

Important: include failed and uncertain item checks even if the answer is mostly good. Do not collapse all items into a global pass.

### Source Context
{source_context}

### Original Task
{task}

### Private Specification Contract Items
{private_contract_items}

### Draft Answer
{draft_answer}
\end{lstlisting}

\paragraph{Repair.}
The repair stage applies one local patch pass using the checker output. It is instructed to preserve correct content and avoid exposing the hidden process.

\begin{lstlisting}[language={},breaklines=true,breakatwhitespace=false,basicstyle=\ttfamily\scriptsize][breaklines=true]
System:
You are a source-grounded patch editor for context learning tasks. Return only the final answer. Do not mention rubrics, private contracts, checkers, repair plans, or hidden evaluation.

User:
Apply one patch pass to the initial answer.

Patch rules:
- Apply the high- and medium-priority repair_plan items unless they conflict with the original task/source.
- Patch locally: preserve the correct structure and wording from the initial answer when possible.
- Make failed or weak requirements visibly satisfied in the final answer; do not rely on implication.
- Remove or qualify unsupported outside facts when source-only discipline is required.
- Keep exact fixed-output formats, JSON, bullet/numbered-list requirements, and role/protocol constraints.
- Do not add meta-commentary, citations, or extra sections unless the task/context requires them.
- If the answer is already a fixed phrase or exact object, output only the corrected fixed phrase/object.

### Source Context
{source_context}

### Original Task
{task}

### Private Specification Contract Items
{private_contract_items}

### Checker JSON
{checker_output_json}

### Initial Answer
{draft_answer}

### Final Answer
\end{lstlisting}

\section{Qualitative Examples of Baseline-to-PSCI Fixes}
\label{app:qualitative_examples}

We provide two qualitative examples showing actual baseline and PSCI outputs side by side on tasks where the baseline produces a substantive, non-empty answer but fails the public rubric. These examples are illustrative only; the quantitative claims in the paper come from the full benchmark evaluation, ablations, and blind human audit. We deliberately select non-terse baseline failures (over 1{,}000 characters) so that the contrast reflects specification acquisition rather than recovery of blank or near-empty outputs.

\subsection{Manufacturing Assembly: Following the Local Runbook}
\label{app:example_assembly}

\paragraph{Task.}
The model must assemble unit \texttt{FR-7B-114} from a multi-turn transcript and eight artifacts. The context defines a 12-step workflow, two conflicting sign-off slips (Rev A at \(15~\mathrm{ft\cdot lb}\), Rev B at \(18~\mathrm{N\cdot m}\)), an ECN superseding prior torque directives, a spacer-count anomaly, an offered impact driver, and a Validation suggestion to skip the mandatory hand spin test. The required output has four parts: a step plan, a blockers/resolutions section, a Spin Check note to Validation, and a Forklift pickup notification.

\paragraph{Scores and selection criterion.}
Baseline: 10/19 rubric items passed, task failed. PSCI: 19/19, task passed. Baseline response: 7{,}264 characters. The baseline is detailed and operationally plausible---this is not a blank-output rescue.

\paragraph{Failed rubric items.}
The baseline missed nine rubric items, of which seven are workflow-ordering and tool-convention specifications:
\begin{itemize}
\itemsep0em
\item ``Does the model secure the assembly with fasteners as \emph{Step 7}? Fail if securing the assembly is out of sequence.''
\item ``Does the model perform the mandatory basic spin test by hand as \emph{Step 8}?''
\item ``Does the model implement the blueprint check and correction as \emph{Step 10} if wobble persists?''
\item ``If the exact wrench is unavailable, does the model specify using the next size up \emph{and applying extra torque}?''
\item Three additional step-ordering items on Steps 1, 3, and 4.
\end{itemize}

\paragraph{Representative contract items induced by PSCI.}
The induced contract (24 items total; full list omitted for space) includes obligations such as:
\begin{itemize}
\itemsep0em
\item follow the source-defined workflow order from notification through transport request, without skipping or reordering required stages;
\item select Rev B and reject the outdated Rev A directive;
\item defer to ECN-defined torque (\(18~\mathrm{N\cdot m}\)) and ignore the conflicting conversion table;
\item install ceramic hybrid bearings by press-fit only;
\item use the Warehouse-provided spacer rather than bench spares;
\item perform the mandatory hand spin test despite Validation's suggestion to skip;
\item if the exact wrench is unavailable, use the next size up \emph{and} apply extra torque per workaround protocol.
\end{itemize}

\paragraph{Baseline output excerpt (verbatim).}
The baseline inserts an extra ``Position spacer S-12'' step, shifting all subsequent steps by one, and accepts the trainee's impact driver:
\begin{lstlisting}[language={},breaklines=true,breakatwhitespace=false,basicstyle=\ttfamily\scriptsize]
5. Install bearings on rotor (press-fit only)
6. Position spacer S-12 correctly
7. Position housing and align mounting holes
8. Secure assembly with fasteners (torque to 18 N.m)
   Drop in the 8 M8 fasteners, finger-tight first.
   Trainee offers impact driver:
      Use impact driver just to snug the bolts lightly in a crisscross pattern at low setting---no final torque with impact.
   If the exact wrench size is "lost in the chaos":
      Apply workaround protocol: grab the next size up wrench/socket to snug if needed...
9. Perform basic spin test by hand (mandatory)
\end{lstlisting}

\paragraph{PSCI output excerpt (verbatim).}
PSCI preserves the source-defined workflow order, explicitly rejects the impact driver, and ties the wrench workaround to extra torque as the protocol requires:
\begin{lstlisting}[language={},breaklines=true,breakatwhitespace=false,basicstyle=\ttfamily\scriptsize]
Step 5 -- Install bearings on rotor (press-fit only)
Step 6 -- Position housing and align mounting holes (spacer S-12 installed within this step per drawing)
Step 7 -- Secure assembly with fasteners
   If correct size socket/wrench isn't found, I'll use the next size up and apply extra torque as specified by our workaround...
   I do not accept the trainee's impact driver offer (Turn 9): Impact risks messing up alignment and torque accuracy. My established workaround is "next size up wrench + extra torque," not "buzz it with an impact."
Step 8 -- Perform basic spin test by hand (mandatory)
Step 10 -- If wobble persists after shakes: check blueprint and correct assembly
\end{lstlisting}

\paragraph{Analysis.}
The baseline is not careless; it produces a detailed plan with most of the correct content (Rev B, \(18~\mathrm{N\cdot m}\), press-fit only, mandatory spin test). It fails because it inserts an extra step, accepts the impact driver, and omits the ``extra torque'' clause from the wrench workaround. These are obligations defined in local documentation that are absent from the user query but explicit in the source. PSCI's induced contract surfaces them before generation, so the final response conforms to the local runbook rather than producing a generic-but-plausible assembly plan.

\subsection{Rowing Synchronization: Producing the Required Analysis Object}
\label{app:example_rowing}

\paragraph{Task.}
The model must produce a rowing synchronization analysis from a multi-artifact context defining a Normalized Dataset v2 (authoritative when artifacts disagree), a sync-score formula, a 90\% efficiency-loss threshold, an excerpt window (strokes 361--368), and a required four-section output (\texttt{<analysis>}, \texttt{<synchronization\_report>}, \texttt{<to\_insights\_agent>}, \texttt{<to\_dashboard\_agent>}).

\paragraph{Scores and selection criterion.}
Baseline: 14/21 rubric items passed, task failed. PSCI: 21/21, task passed. Baseline response: 6{,}051 characters; it performs real quantitative analysis with seat-level offsets, problem pairs, rhythm metrics, and downstream sections---this is not a blank-output rescue.

\paragraph{Failed rubric items.}
The baseline missed seven rubric items, all of which evaluate formula compliance with the locally defined analysis standard:
\begin{itemize}
\itemsep0em
\item ``Does the model correctly calculate sync score using the specified formula?''
\item ``Does the model correctly calculate rhythm consistency?''
\item ``Does the model correctly calculate efficiency loss for \texttt{sync\_score < 90}? Fail if efficiency loss formula is incorrect or threshold is wrong.''
\item ``Does the model calculate time impact correctly for racing scenarios?''
\item ``Does the model calculate wasted power correctly?''
\item ``Does the model escalate unusual patterns to Insights Agent?''
\item ``Does the model refrain from determining \emph{why} a seat is off-time?''
\end{itemize}

\paragraph{Representative contract items induced by PSCI.}
The induced contract includes selected obligations:
\begin{itemize}
\itemsep0em
\item use the four source-specified top-level tags in the required order;
\item treat Normalized Dataset v2 as authoritative;
\item compute offsets relative to seat 1 with millisecond precision;
\item compute sync score using the source-defined formula combining mean absolute catch offset, mean standard deviation, and outlier count;
\item apply the efficiency-loss model \emph{only} when \texttt{sync\_score < 90};
\item avoid causal diagnoses or drill prescriptions in this agent's sections.
\end{itemize}

\paragraph{Baseline output excerpt (verbatim).}
The baseline reports a sync score and applies a proportional efficiency-loss model with no threshold; it does not show the score formula:
\begin{lstlisting}[language={},breaklines=true,breakatwhitespace=false,basicstyle=\ttfamily\scriptsize]
Crew Sync Score: 87% (good)
Estimated Efficiency Loss: 0.9%
...
Impact:
   Estimated time loss in 2k race: ~3.2 seconds (0.9% of 360s)
   Wasted power: ~15 watts of crew power (0.9% of 1680W)
\end{lstlisting}

\paragraph{PSCI output excerpt (verbatim).}
PSCI applies the score formula explicitly and respects the 90\% threshold on the loss model:
\begin{lstlisting}[language={},breaklines=true,breakatwhitespace=false,basicstyle=\ttfamily\scriptsize]
Crew-wide Catch Timing (for sync score)
   Average absolute catch offset (seats 2--8): 15.9 ms
   Average catch-offset std dev (seats 2--8): 1.3 ms
   Seats with |avg catch offset| > 50 ms: 0
   Applied to the model:
   sync_score = 100 - (15.9 * 0.5) - (1.3 * 0.3) - (0 * 5) ~= 91.7 -> 92% (good)
Estimated Efficiency Loss: 0.0% (sync above 90% threshold; model indicates no material sync-driven loss)
...
Estimated time loss in a 2k race: With 0.0% efficiency loss, projected sync-driven time loss ~= 0.0 s
Estimated wasted power: With 0.0% efficiency loss, projected wasted power ~= 0 W
\end{lstlisting}

\paragraph{Analysis.}
Both answers analyze synchronization in detail and produce all four required sections. The baseline fails because its impact calculations cascade from an incorrect application of the loss model: it treats efficiency loss as proportional to sync deficit rather than gated on the source-defined 90\% threshold, so its 2k time loss and wasted power figures are also wrong. PSCI succeeds because the induced contract makes the threshold explicit before generation, and PSCI's output shows the formula being applied, which the baseline does not. The contrast illustrates that specification acquisition matters even when content analysis is largely present: a quantitatively rich answer can still miss the local definition of a valid calculation.

\paragraph{Summary across both examples.}
In both cases the baseline produces a long, substantive, locally plausible answer that gets most of the content right. It fails on obligations that are stated in the source context but absent from the user query: source-defined workflow order, the conjunction in a tool-substitution rule, the threshold on a loss model, and the source-defined formula for a derived metric. PSCI does not merely generate more content; it generates content that satisfies the local specification surface, which is what CL-Bench's strict rubric measures.

\end{document}